\documentclass{article}

\PassOptionsToPackage{numbers, compress}{natbib}

\usepackage[final]{neurips_2024}

\usepackage[utf8]{inputenc} % allow utf-8 input
\usepackage[T1]{fontenc}    % use 8-bit T1 fonts
\usepackage{hyperref}       % hyperlinks

\usepackage{array, multirow}
\usepackage{arydshln}
\usepackage{booktabs}

\usepackage{url}            % simple URL typesetting
\usepackage{amsfonts}       % blackboard math symbols
\usepackage{nicefrac}       % compact symbols for 1/2, etc.
\usepackage{microtype}      % microtypography
\usepackage{xcolor}         % colors

\usepackage{times}
\usepackage{latexsym}

\usepackage{inconsolata}

\usepackage{graphicx}

\usepackage{algorithm, algpseudocode}
\usepackage{amsmath}

\title{Post Training Quantization of Large Language Models \\with Microscaling Formats}

\author{%
  Sayeh Sharify \\
  d-Matrix\\
  Santa Clara, CA, USA\\
  \texttt{sayehs@d-matrix.ai} \\
  \And
  Utkarsh Saxena \thanks{Work done when the author was an intern at d-Matrix.}\\
  Purdue University \\
  West Lafayette, IN, USA\\
  \texttt{saxenau@purdue.edu} \\
  \And
  Zifei Xu \\
  d-Matrix\\
  Santa Clara, CA, USA\\
  \texttt{xuzifei@d-matrix.ai} \\
  \AND
  Wanzin Yazar \\
  d-Matrix\\
  Santa Clara, CA, USA\\
  \texttt{wyazar@d-matrix.ai} \\ 
  \And
  Ilya Soloveychik \\
  d-Matrix\\
  Santa Clara, CA, USA\\
  \texttt{ilyas@d-matrix.ai} \\
  \And
  Xin Wang  \\
  d-Matrix\\
  Santa Clara, CA, USA\\
  \texttt{xwang@d-matrix.ai} \\
}

\begin{document}

\vspace*{-2mm}
\maketitle
\vspace*{-2mm}
\begin{abstract}
  Large Language Models (LLMs) have distinguished themselves with outstanding performance in complex language modeling tasks, yet they come with significant computational and storage challenges. This paper explores the potential of quantization to mitigate these challenges. We systematically study the combined application of three well-known post-training techniques, SmoothQuant, AWQ, and GPTQ, and provide a comprehensive analysis of their interactions and implications for advancing LLM quantization. We enhance the versatility of these methods by enabling quantization to microscaling (MX) formats, extending the applicability of these PTQ algorithms beyond their original fixed-point format targets. We show that combining different PTQ methods enables us to quantize models to 4-bit weights and 8-bit activations using the MXINT format with negligible accuracy loss compared to the uncompressed baseline.
\end{abstract}
\vspace*{-2mm}

\section{Introduction}
\vspace*{-2mm}
\label{sec:Intro}
Large Language Models (LLMs) have emerged as extremely powerful tools to process and generate natural language. However, their high computational demand and energy consumption make widespread adoption of these models in everyday tasks challenging. One way to address these challenges is post-training quantization, a technique that involves reducing the precision of model parameters and/or activations from the original bit-width to formats with fewer bits. Quantization can significantly reduce the memory footprint and computational requirements of these models, making them more accessible and easily deployable on a wider range of hardware, including mobile and edge devices. However, when applying quantization, outliers with large magnitudes remain an open challenge because they stretch the quantization range, leaving fewer effective bits available for the majority of values, which leads to significant quantization errors and accuracy degradation~\cite{dettmers2022llm}. To address this issue, Xiao et al. proposed SmoothQuant~\cite{xiao2023smoothquant}, a quantization technique that smooths out the activation outliers by migrating the quantization difficulty from activations to weights with a provably equivalent transformation. Lin et al., proposed AWQ~\cite{lin2023awq}, a weight only quantization algorithm that mitigates the quantization error by channel-wise scaling of the salient weights~\cite{lin2023awq}. Similarly, Frantar et al. proposed GPTQ~\cite{frantar2022gptq}, a scalable one-shot quantization method that utilizes Frobenius norm layer-wise loss optimization to quantize weights. In this work, we systematically study the combined application of these three algorithms and provide a comprehensive analysis of their interactions and implications for advancing LLM quantization to various fixed-point and microscaling (MX) formats.

\vspace*{-1mm}
\paragraph{Microscaling format.} 
The microscaling (MX) format for neural net computation was proposed by prior work, first as MSFP~\cite{darvish2020pushing} and later subsumed by an emerging industry standard \emph{microscaling formats}~\cite{rouhani2023microscaling}. 
% Specifically MXINT8, a microscaling format that enables high-accuracy inference using half the memory footprint and twice the throughput of FP16, is an emerging industry standard endorsed by Microsoft, AMD, Arm, Intel, Meta, NVIDIA, and Qualcomm~\cite{rouhani2023microscaling}, already seeing adoption in today's hardware products, such as Qualcomm AI100 Accelerator~\cite{QualcommA100}.
Specifically, MXINT8 is a microscaling format that enables high-accuracy inference using approximately half the memory footprint of FP16. It is an emerging industry standard endorsed by Microsoft, AMD, Arm, Intel, Meta, and NVIDIA~\cite{rouhani2023microscaling} and is already seeing adoption in today's hardware products, such as the Qualcomm cloud AI100 Accelerator~\cite{QualcommA100} and NVIDIA Blackwell GPUs~\cite{NvidiaBlackwell-MX}.

The MX format is a variant of block data representation that uses shared scale factors to represent a group of values. It is characterized by three key components: 1) the scale factor data type, 2) the data type and precision of individual elements, and 3) the scaling block size. The scale factor is uniformly applied across a block of elements. This paper focuses on MX formats that use the \emph{INT} data type for individual elements, referred to as \emph{MXINT}. More details on the microscaling format are provided in Section~\ref{sec:mx} of the appendix.

\vspace*{-2mm}
\paragraph{Notation.} 
Throughout the paper we denote a microscaling (MX) format with scaling block size of $b$, $\textit{8-bit}$ shared scaling factor, and $d$ bits per element by $\textit{\textbf{MXINTd-b}}$. For example, $\textit{MXINT4-128}$ represents an MX format with $4$ bits per element, $8$ bits shared exponent across $128$ values within a block. Similarly, a fixed-point value with $i$ integer bits and no fractional bits is denoted by $\textit{\textbf{INTi}}$. For instance, $\textit{INT4}$ specifies a fixed-point value with $4$ integer bits and no fractional bits.

\vspace*{-2mm}
\paragraph{Contributions.}
\begin{itemize}
\item We adopt SmoothQuant, AWQ, and GPTQ to support quantization to microscaling (MX) data formats, extending their compatibility beyond the originally targeted fixed-point formats in the proposed methods.
\item We study the interaction of SmoothQuant, AWQ, and GPTQ using state-of-the-art models like Llama2 and Llama3.1, offering a comprehensive analysis of their impact on advancing LLM quantization. Our findings demonstrate that the pairs SmoothQuant and GPTQ, as well as AWQ and GPTQ, are synergistic, especially at more restrictive bit-widths.
\end{itemize}

The remainder of this paper is structured as follows: Section~\ref{sec:quant-adaptation} provides a brief overview of three post-training quantization algorithms examined in this study and details their adaptation to support quantization to microscaling data formats. Section~\ref{sec:experiment-setup} describes the experimental setup, presents quantization results from applying these three algorithms both individually and in combination, and includes a Pareto frontier analysis of the quantized models. Section~\ref{sec:relatedwork} reviews related work. Section~\ref{sec:discussion} provides a brief discussion on the empirical findings of this work, while Section~\ref{sec:conclusion} concludes the paper. The limitations of this work are explained in Section~\ref{sec:limitations}.

\vspace*{-2mm}
\section{Quantization algorithms adaptation methodology}
\label{sec:quant-adaptation}
\vspace*{-1mm}
Various Post-Training Quantization (PTQ) techniques have emerged to reduce memory bandwidth requirements during LLM inference by quantizing weights and/or activations to lower precisions while maintaining accuracy. In this work, we examine the interaction of three well-known PTQ algorithms for LLMs: GPTQ~\cite{frantar2022gptq}, SmoothQuant~\cite{xiao2023smoothquant}, and AWQ~\cite{lin2023awq}. GPTQ is a weight-only quantization technique that reduces quantization error by quantizing the weight matrix column-wise, across the channels, and sequentially updating the unquantized weights using second-order statistic of the activations to mitigate the error. SmoothQuant scales both activations and weights to reduce the activation's dynamic range, by transferring some of the quantization challenges from activations to weights. AWQ scales weights according to activation magnitudes for improved quantization.  For further details on these three algorithms, please refer to Section~\ref{sec:ptq} of the appendix. The remainder of this section details the generalization of GPTQ, AWQ, and SmoothQuant to support microscaling (MX) quantization, extending their compatibility beyond the originally targeted fixed-point formats in the initially proposed methods.

\begin{algorithm}[t]
    \caption{Enhanced GPTQ: Quantize $\textbf{W}$ given inverse Hessian $\textbf{H}^{-1} = (2\textbf{X}\textbf{X}^T+\lambda \textbf{I})^{-1}$, GPTQ block size $b_1$, and micro-block size $b_2$.}
    \label{alg:example}
    \begin{algorithmic}[1]
        \State {Input:} $\textbf{W}$, \textbf{H}$^{-1}$     \hspace*{28mm} \textcolor{gray}{// Weight and Hessian inverse matrices}
        \State {Input:} $d_{row}$, $d_{col}$ \hspace*{26mm} \textcolor{gray}{// Row and Column dimensions of W}
        \State {Input:} $b_1$, $b_2$     \hspace*{32.5mm} \textcolor{gray}{// GPTQ block size and Micro-block size}
        \State {Variable:} $\textbf{E}$    \hspace*{34mm} \textcolor{gray}{// Quantization error matrix}
        \State {Output:} $\textbf{Q}$      \hspace*{35.5mm} \textcolor{gray}{// Quantized weight matrix}
        
        \State {Initialize:} $\textbf{Q} \gets 0_{d_{row} \times d_{col}}$
        \State {Initialize:} $\textbf{E} \gets 0_{d_{row} \times d_{col}}$
        \State {Initialize:} $\textbf{H}^{-1} \gets Cholesky(\textbf{H}^{-1})^T$

        \For{$i=0, b_1, 2b_1, ...$} 
            \For{$j=i, i+b_2, i+2b_2,... , i+b_1-1$}
                \State $k \gets j+b_2$  \hspace*{23mm} \textcolor{gray}{// helper index}
                \State $\textbf{Q}_{:,j:k} \gets quant(\textbf{W}_{:,j:k})$
                \State $\textbf{E}_{:,j:k} \gets (\textbf{W}_{:,j:k} - \textbf{Q}_{:,j:k})([\textbf{H}^{-1}]_{j:k,j:k})^{-1}$
                \State $\textbf{W}_{:,k:} \gets \textbf{W}_{:,k:} - \textbf{E}_{:,j:k} [\textbf{H}^{-1}]_{j:k, k:}$
            \EndFor
            \State $\textbf{W}_{:,i+b_1:} \gets \textbf{W}_{:,i+b_1:} - \textbf{E}_{:,i:i+b_1} [\textbf{H}^{-1}]_{i:i+b_1, i+b_1:}$
        \EndFor
        \State {Return:} \textbf{Q}
    \end{algorithmic}
    \vspace*{-1mm}
\end{algorithm}

\subsection{GPTQ adaptation to MX format} 
\vspace*{-1mm}
To make GPTQ compatible with the MX format, we modify the algorithm to quantize and update weight values block-wise instead of the originally proposed column-wise updates. Algorithm~\ref{alg:example} illustrates the quantization procedure: The weight matrix is divided into blocks (Line 4: $b_1$), which are further subdivided into micro-blocks (Line 5: $b_2$). Blocks of consecutive micro-blocks are quantized at each step using inverse Hessian information stored in the Cholesky decomposition (Lines 13-18), and the remaining weights are updated at the end of the step (Line 19). This quantization process is applied recursively to different consecutive weight blocks until the entire weight matrix is quantized (Line 12). Note that for quantizing weight matrix to a specific MX format, the micro-block size in the algorithm, $b_2$, should be a multiple of the block size of the MX format. For more details on the GPTQ algorithm please refer to Section~\ref{sec:gptq} of the appendix.

\vspace*{-3mm}
\subsection{SmoothQuant and AWQ adaptation to MX format}
\vspace*{-2mm}
For quantization to the MX format using SmoothQuant and AWQ, we directly calculate per-channel scaling factors to mitigate outliers in activations and/or weights, similar to the approaches proposed in the original paper, and skip the additional calibration phase required for quantization to fixed-point formats~\cite{xiao2023smoothquant, lin2023awq}. 
Sections~\ref{sec:smoothquant} and~\ref{sec:awq} of the appendix provide more details on the SmoothQuant and AWQ algorithms, respectively.

\section{Challenges in studying PTQ algorithms interactions}
\vspace*{-2mm}
This section highlights the challenges encountered when applying the post-training quantization algorithms studied. We found that some algorithms are incompatible, and for those that are compatible, the order of application is crucial. For instance, both AWQ and SmoothQuant aim to moderate the dynamic range of weight values by calculating scaling factors based on activation and weight tensors. However, despite using different formulas to calculate these scaling factors, we did not observe any benefit from combining the two algorithms. In contrast, GPTQ paired with either AWQ or SmoothQuant proved to be synergistic. When combining GPTQ with SmoothQuant or AWQ, it is essential to first smooth the weight range using SmoothQuant or AWQ, then apply GPTQ to the smoothed weights. Reversing this order results in a significant performance degradation. Section~\ref{sec:res} provides more details on the quantization results using different combinations of these post training quantization algorithms.

\section{Experiments}
\label{sec:experiment-setup}
\vspace*{-2mm}
\subsection{Setup.} 
\vspace*{-1mm}
We evaluate the impact of the SmoothQuant, AWQ, and GPTQ techniques on quantization of the Llama2, the Llama3.1, and the Qwen2 family. We employ various fixed-point and MXINT formats with different bit-widths for our assessment and report the perplexity of the quantized models on \textit{WikiText-2}~\cite{merity2016pointer} and \textit{C4}~\cite{raffel2020c4}. We also measure the accuracy of the quantized models on eight zero-shot commonsense reasoning
tasks. (See Section~\ref{par:datasets}.) Moreover, we study the impact of applying GPTQ, SmoothQuant, and AWQ individually, as well as the combined effects of GPTQ with AWQ and GPTQ with SmoothQuant. The following sections provide a detailed explanation of the experimental setup.

\vspace*{-2mm}
\paragraph{Models.}
We evaluated various quantization methods using the Llama2~\cite{touvron2023llama2}, the Llama3.1~\cite{llama3.1meta}, and Qwen2~\cite{yang2024qwen2} families. These LLMs are widely accepted in the machine learning community for their superior performance compared to other open-source LLMs~\cite{dettmers2022llm, frantar2022gptq, xiao2023smoothquant, lin2023awq, li2024llava, adler2024nemotron}. Llama also serves as the foundation for many popular open-source models such as Alpaca~\cite{taori2023alpaca}, Vicuna~\cite{chiang2023vicuna}, and Stable Beluga~\cite{StableBeluga}. For the Llama2 family, unless specified otherwise, we used the default maximum sequence length of 4,096 that was employed during pre-training. For the Llama3.1 and Qwen2 families, we used a reduced maximum sequence length of 8,096 instead of the original 128k used in pre-training, due to limitations in available GPU memory.

\vspace*{-2mm}
\paragraph{Datasets and tasks.}
\label{par:datasets}
Following previous work ~\cite{dettmers2022llm, xiao2023smoothquant, frantar2022gptq, lin2023awq, dettmers2023case, yao2022zeroquant}, we measured the perplexity of quantized language models on \textit{WikiText-2}~\cite{merity2016pointer}, and \textit{C4}~\cite{raffel2020c4} as perplexity reliably reflects the performance of LLMs~\cite{dettmers2023case, lin2023awq}. Unless otherwise stated, the \textit{test} split of the dataset is used to evaluate the models. 
Moreover, to assess the performance of the quantized models on downstream tasks, we measured their accuracy on eight zero-shot commonsense reasoning tasks, including ARC-easy, ARC-challenge~\cite{clark2018arc}, BoolQ~\cite{clark2019boolq}, PIQA~\cite{bisk2020piqa}, SIQA~\cite{sap2019socialiqa}, HellaSwag~\cite{zellers2019hellaswag}, OBQA~\cite{obqa2018can}, and WinoGrande~\cite{sakaguchi2021winogrande}. 

\vspace*{-2mm}
\paragraph{Quantization formats.} 
We evaluated models using different microscaling and fixed-point quantization formats. For the fixed-point quantization, we calibrated the models using $128$ random input sentences from \textit{WikiText-2-train} to estimate the dynamic range of activations. We utilized \mbox{\emph{MinMaxObserver}} to find the range of activations, and calculated the zero-point and the scale parameters for the activations and weights in per-channel granularity levels. For the MXINT format, unless otherwise specified, the blocking dimension of a given tensor is the last dimension.

\vspace*{-2mm}
\paragraph{Activation smoothing.}
For SmoothQuant, we calculated the per-channel scaling factor for activations and weights using the formula stated in Equation~\ref{eq:smooth}. As in the previous work~\cite{xiao2023smoothquant, lee2023enhancing}, we consistently use a migration strength ($\alpha$) value of $0.5$ across all models throughout the paper. To calculate the scaling factors, we gathered the statistics of activations using $128$ random sentences from the \textit{WikiText-2-train} dataset. Once we calculated the scaling factors, we used the same values to evaluate the models with different quantization formats.  

\vspace*{-2mm}
\paragraph{Targeted layers.}
Similar to the previous work~\cite{xiao2023smoothquant}, we apply smoothing on the input activation of the self-attention and the feed-forward layers of LLMs. Unless stated otherwise, we transform all \textit{Linear} layers to the specified quantization format while keeping the activation/weight in the original format for other layers including \textit{GELU}, \textit{Softmax}, and \textit{LayerNorm}. 

\vspace*{-2mm}
\paragraph{AWQ setup.} For quantization with AWQ, we use 128 examples of sequence length 512 from \textit{WikiText-2-train} as the calibration dataset. Following \cite{lin2023awq}, we find optimal scales and clipping values by performing a grid search. To obtain best scaling values, we define the search space as $\mathbf{s} = \mathbf{s}_{\mathbf{x}}^{\alpha}$, where $\mathbf{s}_{\mathbf{x}}$ is the average magnitude of activations, and $\alpha^* = \arg\min_{\alpha} \mathcal{L}(\mathbf{s}_{\mathbf{x}}^{\alpha})$. We perform a grid search over the interval [0,1) to find the best $\alpha$. The optimal clipping value is also determined by minimizing the mean squared error (MSE) of the quantization.

\subsection{Main results}
\label{sec:res}

\begin{table*}[t!] 
\centering
\footnotesize
\caption{Perplexity score on \textit{WikiText-2-test} and averaged accuracy on eight zero-shot common sense reasoning tasks for the Llama2-7B, Llama2-13B, and Llama3.1-8B models, when quantized to fixed-point and MXINT formats using different post-training quantization techniques. $0\texttt{-}shot^8$ includes ARC-challenge, ARC-easy, BoolQ, HellaSwag, OBQA, PIQA, SIQA, and WinoGrande tasks. A, W, SQ, and RTN denote activation, weight, SmoothQuant, and round to nearest, respectively. We used \textit{per-channel affine} quantization for the fixed-point formats. For the MXINT formats, we used block size of 128. \texttt{+}: GPTQ weight quantization is used, $\uparrow$ higher is better, $\downarrow$: lower is better.}

\begin{tabular}{l|l|l||cc|cc|cc}
    \toprule
          \multirow{3}{*}{Bit-width}& \multirow{3}{*}{Format}    & \multirow{3}{*}{Method}      & \multicolumn{2}{c|}{Llama2-7B} & \multicolumn{2}{c|}{Llama2-13B} & \multicolumn{2}{c}{Llama3.1-8B} \\ \cdashline{4-9}
              &      &        & $0\texttt{-}shot^8$ & Wiki & $0\texttt{-}shot^8$ & Wiki & $0\texttt{-}shot^8$ & Wiki \\
              &      &        & ($\uparrow$) & ($\downarrow$) & ($\uparrow$) & ($\downarrow$) & ($\uparrow$) & ($\downarrow$) \\ \hline \hline
          A:16,W:16  & A:FP16,W:FP16 & N/A  & \textbf{59.89} & \textbf{5.12} & \textbf{62.80} & \textbf{4.57} & \textbf{63.60} & \textbf{5.61}  \\ \hline
                       &             & RTN  & 59.88 & 5.13          & 62.52 & \textbf{4.58} & 63.59 & \textbf{5.62}  \\
                       &             & GPTQ & 59.69 & 5.13          & 62.62 & \textbf{4.58} & 63.60 & \textbf{5.62}  \\
                       &             & SQ   & 59.87 & \textbf{5.12} & 62.72 & \textbf{4.58} & 63.57 & \textbf{5.62}  \\
                       &             & AWQ  & \textbf{59.97} & \textbf{5.12} & \textbf{62.80} & \textbf{4.58} & \textbf{63.66} & \textbf{5.62}  \\
                       & \multirow{-5}{*}{A: MXINT8,} & SQ+ & 59.87 & \textbf{5.12} & - & \textbf{4.58} & 63.57 & \textbf{5.62}  \\ 
                       & \multirow{-5}{*}{W: MXINT8} & AWQ+ & 59.90 & \textbf{5.12} & 62.64 & \textbf{4.58} & 63.56 & \textbf{5.62} \\ \cdashline{2-9}
                       &            & RTN  & 58.54 & \textbf{5.15} & 61.50 & \textbf{4.60} & 62.97 & \textbf{5.69}  \\ 
                       &            & GPTQ & 58.52 & \textbf{5.15} & \textbf{61.61} & \textbf{4.60} & 63.00 & \textbf{5.69}  \\
                       &            & SQ   & 58.69 & \textbf{5.15} & 61.56 & \textbf{4.60} & \textbf{63.25} & \textbf{5.69}  \\
                       &            & AWQ  & 58.23 & 5.17          & 61.58 & 4.62          & 62.72 & 5.76           \\
                       & \multirow{-5}{*}{A: INT8,} & SQ+ & \textbf{58.71} & \textbf{5.15} & - &\textbf{4.60} & 63.14 & \textbf{5.69}  \\
    \multirow{-12}{*}{A:8, W:8} & \multirow{-5}{*}{W: INT8}  & AWQ+ & 58.15 & 5.17 & 61.46 & 4.62 & 62.52 & 5.76  \\ \hline
                       &            & RTN  & \textbf{58.99} & 5.55          & 61.83 & 4.82          & 60.97 & 6.99           \\ 
                       &            & GPTQ & 57.66 & 5.45          & 61.53 & 4.76          & 60.81 & 6.39           \\
                       &            & SQ   & 57.59 & 5.60          & 61.06 & 4.93          & 59.52 & 7.01           \\
                       &            & AWQ  & 58.43 & 5.43          & \textbf{62.11} & 4.77 & \textbf{61.52} & 6.40           \\
                       & \multirow{-5}{*}{A: MXINT8,} & SQ+ & 57.88 & 5.48 & - & 4.84    & 60.09 & 6.50   \\
                       & \multirow{-5}{*}{W: MXINT4} & AWQ+ & 58.21 & \textbf{5.37} & 61.71 & \textbf{4.73} & 61.18 & \textbf{6.21} \\ \cdashline{2-9}
                       &            & RTN  & 56.69 & 5.91          & 60.14 & 4.97          & 59.73 & 8.11           \\ 
                       &            & GPTQ & 56.53 & 5.67          & 59.54 & 4.85          & 56.46 & 12.56          \\
                       &            & SQ   & 55.27 & 6.34          & 57.34 & 5.56          & 54.51 & 8.86           \\
                       &            & AWQ  & \textbf{57.64} & 5.61 & \textbf{60.86} & 4.85 & \textbf{61.89} & 6.94           \\
                       & \multirow{-5}{*}{A: INT8,} & SQ+ & 56.48 & 5.78 & - & 5.12          & 56.30 & 7.27           \\
    \multirow{-12}{*}{A:8, W:4} & \multirow{-5}{*}{W: INT4} & AWQ+ & 57.16 & \textbf{5.53} & 60.12 & \textbf{4.80} & 60.83 & \textbf{6.56}      \\ \hline
                       &            & RTN  & 51.57 & 9.89          & 54.74 & 6.32          & 41.52 & 124.48         \\ 
                       &            & GPTQ & 49.38 & 8.49          & 55.52 & 5.66          & 43.56 & 12.87          \\
                       &            & SQ   & 45.91 & 23.54         & 50.00 & 8.26          & 33.16 & 2088        \\
                       &            & AWQ  & \textbf{53.92} & 7.14 & \textbf{56.92} & 5.93          & \textbf{50.24} & 16.22          \\
                       & \multirow{-5}{*}{A: MXINT8,} & SQ+ & 49.21 & 7.61 & - & 6.21      & 40.94 & 13.82 \\
                       & \multirow{-5}{*}{W: MXINT3} & AWQ+ & 51.03 & \textbf{6.87}    & 55.84 & \textbf{5.53} & 49.13 & \textbf{8.88}  \\ \cdashline{2-9}
                       &            & RTN  & 32.56 & 9752          & 36.55 & 147.7       & 32.70 & 28925       \\ 
                       &            & GPTQ & 37.85 & 73.36         & 46.10 & 8.57         & 32.85 & 525.71         \\
                       &            & SQ   & NaN   & NaN           & 32.91 & 9305      & 34.20 & 35424     \\
                       &            & AWQ  & 37.15 & 81.52         & 50.83 & 8.08         & 39.18 & 152.94          \\
                       & \multirow{-5}{*}{A: INT8,} & SQ+ & NaN & NaN          & - & 53.36       & 32.64 & 2861 \\
    \multirow{-12}{*}{A:8, W:3} & \multirow{-5}{*}{W: INT3} & AWQ+ & \textbf{44.02} & \textbf{10.18} & \textbf{51.78} & \textbf{6.23}    & \textbf{41.56} & \textbf{27.19}  \\
    \bottomrule

\end{tabular}
\label{tab:llama}
\vspace*{-3mm}
\end{table*}

\begin{table*}[h!] 
\centering
\footnotesize
\caption{Perplexity score on \textit{C4} for the \textit{Qwen2-1.5B} and \textit{Qwen2-7B} models, when quantized to fixed-point and MXINT formats using different post-training quantization techniques. For the C4 dataset, \textit{validation} split of the \textit{realnewslike} subset is used. A, W, and RTN denote activation, weight, and round to nearest, respectively. We used \textit{per-channel affine} quantization for the fixed-point formats. \texttt{+}: GPTQ weight quantization is used. $\uparrow$ higher is better, $\downarrow$: lower is better.}
\begin{tabular}{l|l|l||cc|cc}
    \toprule
          \multirow{3}{*}{Bit-width} & \multirow{3}{*}{Format} & \multirow{3}{*}{Method} & \multicolumn{2}{c|}{Qwen2-1.5B} & \multicolumn{2}{c}{Qwen2-7B}      \\  \cdashline{4-7}
           &      &     & $0\texttt{-}shot^8$ & C4 & $0\texttt{-}shot^8$ & C4      \\
           &      &      & ($\uparrow$) & ($\downarrow$) & ($\uparrow$) & ($\downarrow$) \\ \hline \hline
          A:16, W:16  & A: FP16, W: FP16 & N/A     & \textbf{54.40} & \textbf{12.08} & \textbf{63.45} & \textbf{9.23}  \\ \hline
                       &            & RTN          & \textbf{54.43} & 12.11          & 63.28 & 9.26  \\
                       &            & GPTQ         & 54.32 & 12.11          & 63.17 & 9.25  \\
                       &            & SmoothQuant  & 54.32 & 12.11	         & 63.33 & 9.25  \\
                       &            & AWQ          & 54.22 & \textbf{12.10} & 63.25 & \textbf{9.24}  \\
                       & \multirow{-5}{*}{A: MXINT8-128,} & SmoothQuant+ & 54.18 & \textbf{12.10} & 63.21 & 9.25  \\
                       & \multirow{-5}{*}{W: MXINT8-128} & AWQ+ & 54.30 & \textbf{12.10} & \textbf{63.35} & \textbf{9.24} \\\cdashline{2-7}
                       &            & RTN          & 54.32 & 12.22          & 62.79 & \textbf{9.37} \\ 
                       &            & GPTQ         & 54.37 & \textbf{12.21} & 62.99 & \textbf{9.37}  \\
                       &            & SmoothQuant  & 54.11 & 12.22          & 63.04 & 9.38  \\
                       &            & AWQ          & 53.69 & 12.24          & 62.67 & 9.39  \\
                       & \multirow{-5}{*}{A: INT8,} & SmoothQuant+ & \textbf{54.46} & \textbf{12.21} & \textbf{63.05} & \textbf{9.37}  \\
    \multirow{-12}{*}{A:8, W:8} & \multirow{-5}{*}{W: INT8} & AWQ+ & 53.62 & 12.24   & 62.88 & 9.39  \\ \hline
                       &            & RTN          & 52.35 & 14.19 & 59.28 & 13.87  \\ 
                       &            & GPTQ         & 52.71 & 13.23 & \textbf{62.24} & 9.73   \\
                       &            & SmoothQuant  & 49.33 & 14.72 & 61.42 & 10.54  \\
                       &            & AWQ          & 52.23 & 13.56 & 61.60 & 9.83   \\
                       & \multirow{-5}{*}{A: MXINT8-128,} & SmoothQuant+ & 50.73 & 13.59 & 61.19 & 9.94 \\
                       & \multirow{-5}{*}{W: MXINT4-128} & AWQ+ & \textbf{52.90} & \textbf{13.15} & 62.13 & \textbf{9.63} \\ \cdashline{2-7}
                       &            & RTN          & 51.52 & 14.96 & 51.70 & 46.64  \\ 
                       &            & GPTQ         & 51.38 & 13.51 & 60.88 & 10.12  \\
                       &            & SmoothQuant  & 48.89 & 16.83 & 53.82 & 24.93  \\
                       &            & AWQ          & \textbf{52.03} & 13.78 & \textbf{61.69} & 10.74  \\
                       & \multirow{-5}{*}{A: INT8,} & SmoothQuant+ & 50.81 & 14.47 & 57.42 & 11.00  \\
    \multirow{-12}{*}{A:8, W:4} & \multirow{-5}{*}{W: INT4} & AWQ+ & 51.52 & \textbf{13.43} & 61.26 & \textbf{9.99} \\ \hline
                       &            & RTN          & 40.87 & 49.70  & 42.82 & 302.36 \\ 
                       &            & GPTQ         & 42.60 & 20.54  & 53.09 & 12.21  \\
                       &            & SmoothQuant  & 37.11 & 204.84 & 45.46 & 29.39  \\
                       &            & AWQ          & \textbf{45.82} & 30.98  & 56.35 & 13.54  \\
                       & \multirow{-5}{*}{A: MXINT8-128,} & SmoothQuant+ & 37.98 & 29.71 & 48.17 & 13.47 \\
                       & \multirow{-5}{*}{W: MXINT3-128} & AWQ+ & 43.60 & \textbf{18.42} & \textbf{58.46} & \textbf{11.26} \\ \cdashline{2-7}
                       &            & RTN          & 34.53 & 1655  & 34.41 & 42055896 \\ 
                       &            & GPTQ         & 37.16 & 34.76 & 39.03 & 25.42    \\
                       &            & SmoothQuant  & 32.94 & 23048 & 33.26 & 2096039  \\
                       &            & AWQ          & \textbf{43.87} & 37.99 & 47.51 & 40.50    \\
                       & \multirow{-5}{*}{A: INT8,} & SmoothQuant+ & 33.36 & 509.23 & 33.06 & 591.29  \\
    \multirow{-12}{*}{A:8, W:3} & \multirow{-5}{*}{W: INT3} & AWQ+ & 43.70 & \textbf{21.13} & \textbf{49.43} & \textbf{16.78}  \\
    \bottomrule

\end{tabular}
\label{tab:qwen2}
\vspace*{-3mm}
\end{table*}

\vspace*{-1mm}
\paragraph{Perplexity.} Table~\ref{tab:llama} illustrates perplexity of the quantized \emph{Llama} models~\cite{touvron2023llama2, llama3.1meta} with three different sizes on \emph{WikiText-2-test} using various MX and fixed-point formats. For all three models, aggressive quantization to small bit-widths penalizes the model performance, while quantizing to higher bit-widths has negligible effect on perplexity. For example, quantizing \mbox{\textit{Llama3.1-8B}} to \textit{MXINT8} preserves the baseline perplexity while quantizing to \textit{MXINT4} increases perplexity by $25\%$ to $6.99$. Moreover, quantization results using different MXINT format delivers better perplexity compared to the fixed-point formats with the same bit-width. For instance, quantizing \textit{Llama2-7B} to \textit{INT4} increases perplexity to $5.91$. Enabling AWQ, and GPTQ jointly, reduces it to $5.53$, while using \textit{MXINT4} and enabling AWQ and GPTQ we can achieve perplexity of $5.37$. Additionally, we found that in all cases except for the quantization of both activations and weights to INT8, AWQ shows superior results compared to SmoothQuant. For the studied models and quantization formats, both SmoothQuant and GPTQ, as well as AWQ and GPTQ, are synergistic, an effect most prominent in more aggressive quantizations. For example, quantizing \textit{Llama2-13B} to \textit{MXINT3} results in perplexity scores of $8.26$ with SmoothQuant and $5.93$ with AWQ. Enabling GPTQ improves these scores to $6.21$ and $5.53$, respectively, with the best perplexity achieved by enabling both AWQ and GPTQ together.

\vspace*{-2mm}
\paragraph{Granularity.} To evaluate the impact of block size on MXINT quantization, we conducted experiments on the \emph{Llama2 and Llama3.1}~\cite{touvron2023llama2} models, using MXINT formats with a block size of \emph{16} instead of the default 128. In this experiment, we applied several quantization algorithms, including GPTQ, AWQ, and SmoothQuant, to assess their effects under the smaller block size. Our results indicate trends similar to those observed in previous sections, with MXINT formats continuing to demonstrate strong performance across various bit-widths. Detailed results of this experiment are provided in Table~\ref{tab2:llama} of the appendix.

\vspace*{-2mm}
\paragraph{Generalizability.} To evaluate whether our findings generalize to other models and datasets, we conducted similar experiments on the \emph{Qwen2}~\cite{yang2024qwen2} family, measuring perplexity on the \emph{C4}~\cite{raffel2020c4} dataset. Table~\ref{tab:qwen2} presents the detailed results for the \emph{Qwen2-1.5B} and \emph{Qwen2-7B} models. Our observations align with earlier results from the Llama2, and Llama3.1 models on the WikiText-2 dataset. Specifically, the MXINT format consistently outperforms the INT format at the same bit-width. Additionally, except for the INT8 case, the best results are obtained by enabling both AWQ and GPTQ together, confirming their synergistic effect.

\vspace*{-2mm}
\paragraph{Downstream tasks.}
Similar to previous work~\cite{liu2024spinquant}, we evaluated the accuracy of the quantized models on eight zero-shot commonsense reasoning tasks, including ARC-easy, ARC-challenge~\cite{clark2018arc}, BoolQ~\cite{clark2019boolq}, PIQA~\cite{bisk2020piqa}, SIQA~\cite{sap2019socialiqa}, HellaSwag~\cite{zellers2019hellaswag}, OBQA~\cite{obqa2018can}, and WinoGrande~\cite{sakaguchi2021winogrande}. Tables~\ref{tab:llama}, and~\ref{tab:qwen2} show the average accuracy across these tasks for the Llama2, Llama3.1, and Qwen2 model families. Full results are reported in Tables~\ref{tab:llama2-7b_llama2-13b_eval_harness},~\ref{tab:llama3-8b_eval_harness}, and~\ref{tab:Qwen_eval_harness} of the appendix Section.

Overall, downstream tasks tend to be more tolerant and less error-prone compared to language modeling tasks such as perplexity. Accordingly, for the downstream tasks the performance gap between the baseline and quantized models across all studied networks are smaller. (See Tables~\ref{tab:llama}, and~\ref{tab:qwen2}). For instance, quantizing the \emph{Llama2-7B} model to \emph{MXINT4} yields approximately $99\%$ of the baseline accuracy without the need to enable additional quantization techniques such as AWQ or SmoothQuant (Table~\ref{tab:llama}). This indicates that quantized models can retain high accuracy even at lower bit-widths in downstream tasks.

Despite downstream tasks being less error-prone, in general, our findings are largely consistent with those from the perplexity experiments. Notably, 8-bit quantization does not require additional quantization techniques, with simple rounding to the nearest value closely matching baseline performance. Furthermore, across all models, \emph{MXINT} consistently outperforms \emph{INT} at the same bit-width, with the best accuracy achieved using MXINT formats surpassing that of INT formats. Additionally, except in the case of 8-bit quantization, AWQ consistently surpasses SmoothQuant. However, unlike the observation with the perplexity experiments, when quantizing to 4-bit, AWQ alone is typically sufficient, and adding GPTQ tends to slightly degrade accuracy (Tables~\ref{tab:llama}, and~\ref{tab:qwen2}). That said, in cases of more aggressive quantization to INT3, the best results across all three model families, are still mostly achieved when both AWQ and GPTQ are enabled, highlighting the synergistic effect of combining these techniques with more restrictive quantizations.

\begin{figure*}[t!]
\centering
  \includegraphics[scale=0.51]{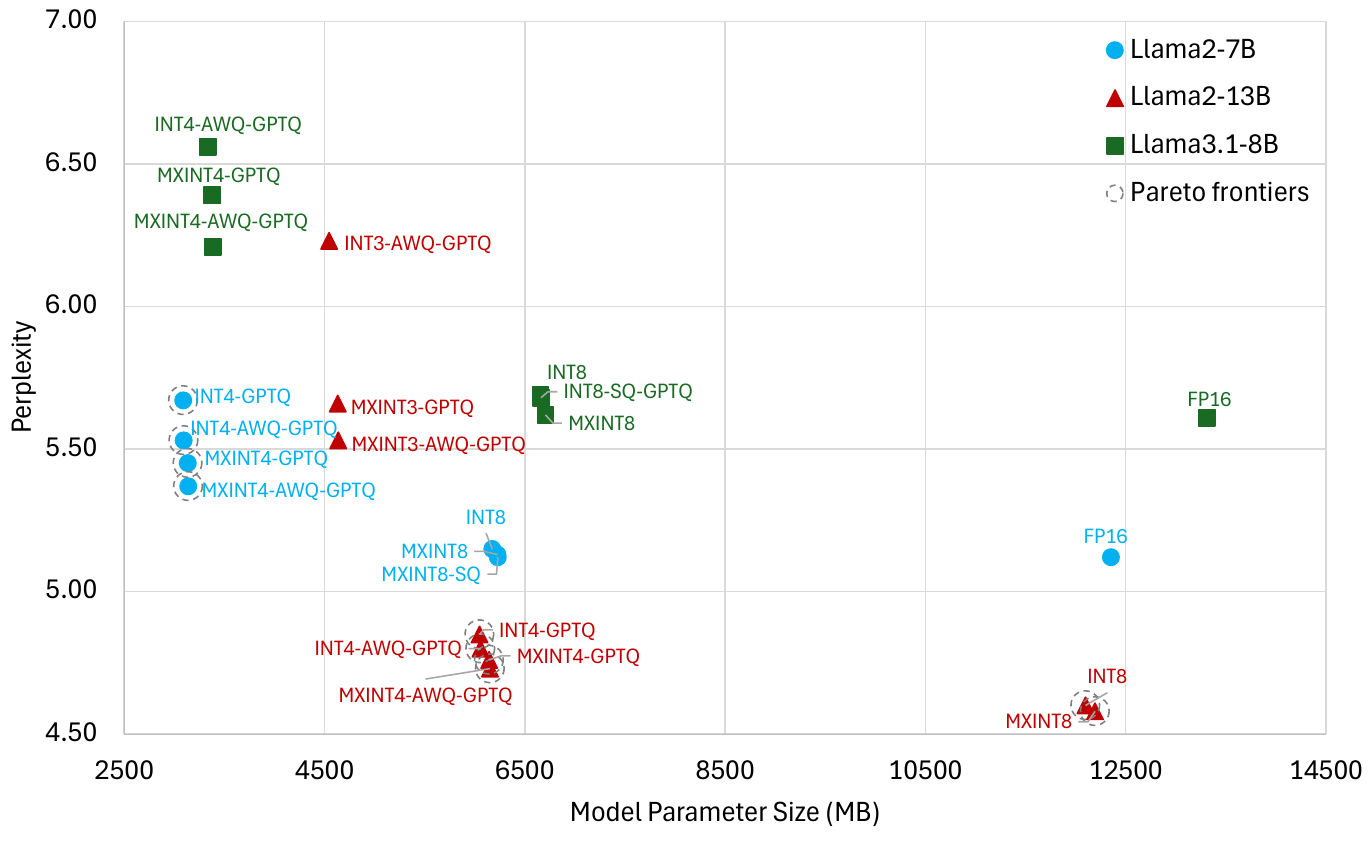} 
    \caption{Perplexity for the LLaMA2 and Llama3.1 families when quantized to 8-bit, 4-bit, and 3-bit MXINT, and INT formats. The Y-axis represents perplexity. The X-axis represents model parameter size including the additional scale parameters required by the SmoothQuant and AWQ quantization methods. Note that the GPTQ algorithm does not introduce any additional model parameters during the inference. \textit{A}, \textit{W}, and \textit{SQ} denote activation, weight, and SmoothQuant. The points corresponding to the quantized models on Pareto frontiers are indicated by a gray dashed circle.}
    \label{fig:LLaMA-ParetoFrontiers}
    \vspace*{-4mm}
\end{figure*}

\vspace*{-2mm}
\subsection{Pareto frontier Study}
\label{sec:Pareto}
\vspace*{-1mm}
The objective of a quantization method is to reduce the model size while preserving its accuracy. In the experiments conducted in this study, the concept of the \emph{Pareto frontier} becomes relevant in determining the most suitable quantization method for each model under a size constraint. A model is deemed to be on the Pareto frontier if no other model exists with both a smaller size and lower perplexity. Figure~\ref{fig:LLaMA-ParetoFrontiers} illustrates perplexity of the Llama2 and Llama3.1 families on WikiText-2-test as a function of model parameter size. Points corresponding to the quantized models on Pareto frontiers are marked with a gray dashed circle. We observe that aggressive quantization to 3-bit significantly penalizes performance, leaving none of the 3-bit points on the Pareto frontier. With more relaxed 4-bit quantization, models appear on the Pareto frontier when either GPTQ is applied (e.g., Llama2-13B, INT4) or when GPTQ is combined with AWQ (e.g., Llama2-7B, MXINT4). Lastly, for 8-bit quantization (e.g., Llama2-13B, MXINT8, INT8), none of GPTQ, AWQ, or SmoothQuant is required to achieve strong performance, as the 8-bit width alone suffices to preserve baseline accuracy effectively.

\section{Related Work}
\label{sec:relatedwork}
\vspace*{-2mm}
\paragraph{Model quantization methods.} Quantization is a technique that lowers the bit precision of deep learning models, effectively reducing model size and accelerating inference. 
There are two primary categories of quantization techniques: Quantization-Aware Training (QAT), which leverages backpropagation to update quantized weights~\cite{bengio2013estimating, choi2018pact, nagel2021white, gholami2022survey, liu2024spinquant}, and Post-Training Quantization (PTQ), which typically requires no additional training. Quantization-aware training methods cannot easily scale up to quantize giant LLMs. Consequently, PTQ methods are commonly employed for quantizing LLMs~\cite{jacob2018quantization, nagel2019data, nagel2020up, wang2020towards, hubara2021accurate, li2021brecq, deng2023mixed}. In this work, we studied the interaction of three PTQ methods, SmoothQuant~\cite{xiao2023smoothquant}, AWQ~\cite{lin2023awq}, and GPTQ~\cite{frantar2022gptq}.

\vspace*{-2mm}
\paragraph{Large Language Model quantization.} 
% With the recent open-source releases of language models like Llama~\cite{touvron2023llama2}, researchers are developing cost-effective quantization methods to compress these models for inference:
With the recent open-source releases of language models like Llama~\cite{touvron2023llama2}, researchers are actively working on developing cost-effective methods to compress these large networks for inference. Various approaches have been suggested to tackle the challenges of quantizing LLMs. ZeroQuant~\cite{yao2022zeroquant} and nuQmm~\cite{park2022nuqmm} employ per-token and group-wise quantization schemes for LLMs, requiring customized CUDA kernels. ZeroQuant further proposes layer-wise knowledge distillation, similar to AdaQuant~\cite{hubara2021accurate}, but the largest evaluated model by both ZeroQuant and nuQmm has 20B parameters. 
LLM.int8() identifies activation outliers in a few feature dimensions as a hindrance to the quantization of larger models, and proposes to preserve those dimensions in higher precision using a mixed INT8/FP16 decomposition~\cite{dettmers2022llm}. However, this implementation results in significant latency overhead, sometimes even slower than FP16 inference. 
Similarly, SpQR~\cite{dettmers2023spqr} and OWQ~\cite{lee2024owq} propose to retain outlier features that are difficult to quantize in full-precision, while AWQ~\cite{lin2023awq} mitigates the quantization error for the outliers using grid-searched channel-wise scaling. Additionally, Outlier Suppression~\cite{wei2022outlier} tackles activation outliers by utilizing non-scaling LayerNorm and token-wise clipping. Despite its success with smaller language models such as BERT~\cite{devlin2018bert} and BART~\cite{lewis2019bart}, it falls short in maintaining accuracy for larger LLMs, while SmoothQuant and GPTQ both preserve the performance of LLMs up to 175B parameters~\cite{xiao2023smoothquant, frantar2022gptq}. Lee et al., explored the combined use of AWQ, SmoothQuant, and GPTQ for quantizing LLMs, focusing solely on fixed-point data types in their study~\cite{lee2023enhancing}. QuaRot is a novel quantization scheme that enables end-to-end 4-bit quantization of LLMs including all weights, activations, and the KV cache. By utilizing Hadamard matrices, QuaRot effectively rotates LLMs to eliminate outliers in the activations and KV cache of pre-trained models~\cite{ashkboos2024quarot}. Trukhanov et al., proposed a technique for quantizing KV-cache to low-precision Block Floating-Point (BFP) formats without compromising the resulting model accuracy~\cite{trukhanov2024accurate}.

\vspace*{-2mm}
\section{Discussion}
\label{sec:discussion}
\vspace*{-1mm}
In this section, we discuss the synergistic effects of AWQ/SmoothQuant and GPTQ, as well as the superiority of AWQ over SmoothQuant in quantization to smaller bit-widths (e.g., 4 bits). These insights are based on an initial hypothesis, and further investigation is needed to qualitatively analyze these effects. We defer a more in-depth analysis to future work.

\vspace*{-2mm}
\paragraph{Synergistic effect of PTQ algorithms.} GPTQ quantizes the weights of different layers to lower bit-widths by minimizing the mean squared error associated with the weight values. It achieves this by finding a compressed version of each weight that minimizes the quantization error, ensuring efficient compression with minimal impact on accuracy. However, when used alone, GPTQ may not fully address the interaction between quantized weights and activations, particularly in cases where activations vary significantly due to outliers, potentially leading to accuracy degradation. AWQ and SmoothQuant, on the other hand, ensure that weights are quantized in a way that takes into account how they will impact the activations. By aligning weight quantization with activation behavior, they minimize errors more efficiently. Based on our empirical results (Sections~\ref{sec:res}, and~\ref{sec:Pareto}), we observed that both AWQ and GPTQ, as well as SmoothQuant and GPTQ, exhibit synergistic effects. This synergy enables more aggressive quantization (down to 4-bit or lower) with minimal impact on accuracy, making it highly effective for deploying large models in resource-constrained environments such as edge devices or real-time applications.

\vspace*{-2mm}
\paragraph{Superiority of AWQ over SmoothQuant.} Both SmoothQuant and AWQ aim to address outlier features that are challenging to quantize, mitigating quantization errors by introducing additional operations to the network. SmoothQuant achieves this by smoothing out activation outliers, effectively transferring the quantization difficulty from activations to weights. In contrast, AWQ reduces quantization error by applying channel-wise scaling to the most salient weights.
However, in more aggressive quantization scenarios, such as 4-bit quantization, the process becomes increasingly challenging. Transferring the quantization difficulty from activations to weights, as SmoothQuant does, adds extra complexity to the already demanding task of weight quantization. This additional burden can make SmoothQuant less effective at such low bit-widths. AWQ, by directly managing the quantization difficulty through careful scaling of weights, proves more advantageous in these cases. Therefore, AWQ demonstrates superior performance under extreme quantization conditions.

\section{Conclusion}
\label{sec:conclusion}
\vspace*{-1mm}
To summarize, we demonstrated that for the studied models, quantizations using different MX formats deliver better perplexity compared to fixed-point formats with the same bit-width when the per-channel affine quantization scheme is employed. Particularly, for quantization to MXINT8, none of GPTQ, AWQ, or SmoothQuant are necessary to preserve the baseline accuracy. Notably, we found that for Llama2 and Llama3.1, when quantized to MX formats, AWQ is superior to SmoothQuant. Moreover, AWQ and GPTQ are synergistic, especially, with more aggressive quantization to 3-bit.

%Additionally, contrary to the results of prior research~\cite{xiao2023smoothquant}, we illustrated that when applying SmoothQuant, it suffices to apply only the \mbox{A-W SmoothQuant}, as opposed to the original findings that recommended both A-W and A-A approaches. 
Throughout the paper, we have shown that by utilizing AWQ, and GPTQ and applying MX formats we can quantize the Llama2 and Llama3.1 models to 4-bit weights and 8-bit activations, with minimal perplexity degradation.

%%%%%%%%%%will not be count toward the page limit.%%%%%%%%%%%%%%%%%%
\section{Limitations}
\label{sec:limitations}
\vspace*{-1mm}
With quantization of LLMs, we make the models accessible to more people, which generally comes with security risks, such as potential misuse for generating harmful content. This highlights the need for further investigation into responsible AI practices. On the technical side, due to space and computational resource constraints, we have only reported results for text generation and commonsense reasoning tasks with the Llama2, Llama3.1, Qwen2 models up to 13B parameters. Further investigation of larger models, and broader datasets/tasks remains for future work.

% Bibliography entries for the entire Anthology, followed by custom entries
\bibliographystyle{plain} % We choose the "plain" reference style
\bibliography{refs}

\newpage
\appendix

\section{Microscaling data format}
\label{sec:mx}
The Microscaling (MX) data format, initially introduced in 2020 as Microsoft Floating Point (MSFP, ~\citep{darvish2020pushing}), has since evolved and gained widespread adoption among leading industry players, including Microsoft, AMD, Intel, Meta, Nvidia, and Qualcomm \cite{rouhani2023microscaling}.

\begin{figure}[h!]
  \centering
  \includegraphics[scale=0.45]{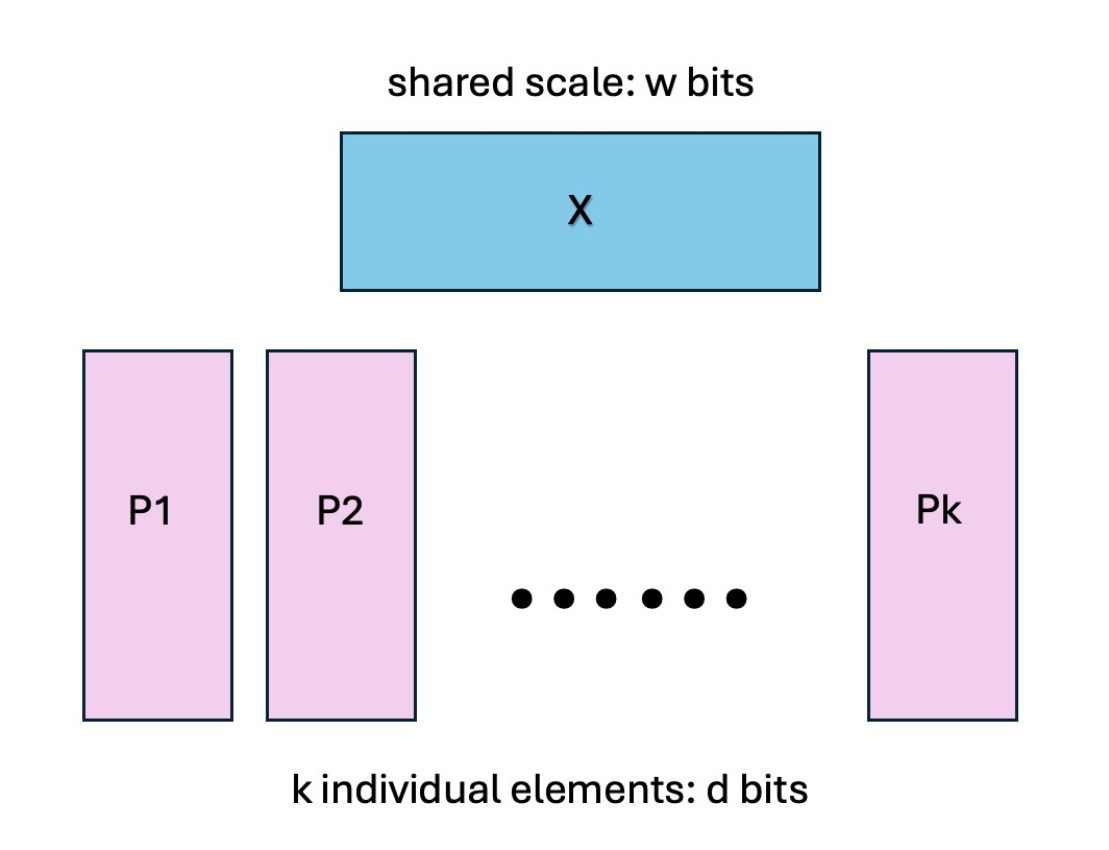} 
  \caption {Illustration of an MX block.}
  \vspace*{-2mm}
  \label{fig:mx}
\end{figure}

The core concept of the MX format is centered around the MX block, where a vector of $k$ numbers share a single scale ($X$) while retaining individual elements ${\{P_i\}^k_{i=1}}$, as shown in Figure~\ref{fig:mx}. The actual value for each of the $k$ numbers in the block can be represented as $v_i = XP_i$ \cite{rouhani2023microscaling}. The data format for the single scale and the data format for individual elements can be independent of each other, while the data format for individual elements needs to be consistent across the $k$ elements in the block \cite{microscaling2023}. An MX block can be represented in $(w+kd)$ bits, where $w$ is the number of bits for shared scale $X$ and $d$ is the number of bits for each individual element. Consequently, the MX format is characterized by three main components:

\vspace*{-2mm}
\begin{enumerate}
\itemsep0em 
  \item Data type of scale $X$
  \item Data type of elements $P_i$
  \item Scaling block size $k$
  \vspace*{-2mm}
\end{enumerate}

The MX format has proven to be highly effective in addressing the challenges of balancing hardware efficiency, model accuracy, and user experience in machine learning applications. According to empirical results, 8-bit MX formats can perform inference directly on FP32 pretrained models with minimal accuracy loss, eliminating the need for additional calibration or finetuning \cite{rouhani2023microscaling}. Furthermore, when using 6-bit MX formats, the inference accuracy remains close to that of FP32 models, especially after applying quantization-aware fine-tuning or post-training quantization methods \cite{rouhani2023microscaling}. Remarkably, the MX format also enables the training of large transformer models using sub-8-bit precision for weights, activations, and gradients, achieving accuracy comparable to FP32 without requiring changes to the training process \cite{rouhani2023microscaling}. 

\section{Post training quantization algorithms}
\label{sec:ptq}

\subsection{GPTQ}
\label{sec:gptq}
GPTQ is a post-training quantization (PTQ) method that uses second-order Hessian information for weight quantization in LLMs~\cite{frantar2022gptq}. It employs layer-wise quantization for each layer $l$ in the network, seeking quantized weights $\hat{\textbf{W}}_l$ that make the outputs $(\hat{\textbf{W}}_l\textbf{X}_l)$ closely approximate those of the original weights $(\textbf{W}_l\textbf{X}_l)$. In other words, GPTQ aims to find~\cite{frantar2022gptq}: 

\begin{equation}
    \texttt{argmin}_{\hat{\textbf{W}}_l} || \textbf{W}_l\textbf{X}_l - \hat{\textbf{W}_l}\textbf{X}_l ||^2_2
    \label{eq:gptq}
\end{equation}

To solve equation~\ref{eq:gptq}, GPTQ quantizes each row of the weight matrix, $\textbf{W}$, independently, focusing on a single weight per row at a time. It consistently updates all not-yet-quantized weights to offset the error introduced by quantizing a single weight. %GPTQ uses second order Hessian information of the weight matrix to quantize the weights. 
Since the objective function in equation~\ref{eq:gptq} is quadratic, its Hessian $\textbf{H}$ can be calculated using the following formula, where $F$ denotes the set of remaining full-precision weights:

\begin{equation}
    \textbf{H}_F = 2\textbf{X}_F\textbf{X}_F^T
    \label{eq:quadratic-hessian}
\end{equation}

Given $\textbf{H}$, the next to be quantized weight, $w_q$, and the corresponding update of all remaining weights in $F$, $\delta_F$, are given by the following formulas, where $\texttt{quant}(w)$ rounds $w$ to the nearest quantized value ~\cite{frantar2022gptq}:

\begin{equation}
    \begin{split}
        w_q = \texttt{argmin}_{w_q} \frac{(w_q - \texttt{quant}(w_q))^2}{[\textbf{H}^{-1}_{F}]_{qq}} \\
        \delta_q = - \frac{w_q - \texttt{quant}(w_q)}{[\textbf{H}^{-1}_{F}]_{qq}} . (\textbf{H}^{-1}_{F})_{:,q}
        \label{eq:wq-dq}
    \end{split}
\end{equation}

For all rows of $\textbf{W}$, GPTQ quantizes weights in the same order. This accelerates the process, as certain computations need to be performed only once for each column rather than once for each weight. Additionally, the vectorized implementation of GPTQ enables processing multiple rows of $\textbf{W}$ simultaneously. For more details on the GPTQ algorithm refer to Frantar et al.’s work~\cite{frantar2022gptq}.

\subsection{SmoothQuant}
\label{sec:smoothquant}
SmoothQuant (SQ) is a quantization method that targets both activations and weights of a model~\cite{xiao2023smoothquant}. In this approach, the activation of a linear layer is scaled by a per-channel smoothing factor $s \in R^{C_i}$ to minimize quantization errors. Simultaneously, the weight of the layer is adjusted in the opposite direction to maintain the mathematical equivalence of the linear layer:

\begin{equation}
    \textbf{Y} = (\textbf{X}\texttt{diag}(s)^{-1})\cdot(\texttt{diag}(s)\textbf{W}) = \hat{\textbf{X}}\hat{\textbf{W}}
    \label{eq:smooth}
\end{equation}

In Equation~\ref{eq:smooth}, $\textbf{X}$ is the original input activation with outliers, and $\hat{\textbf{X}}=\textbf{X}\texttt{diag}(s)^{-1}$ is the smoothed activation. To minimize the quantization error of the input activation, the smoothing factor is selected such that all channels of the smoothed input activation have the same maximum magnitude. Accordingly, $s$ is set to:

\begin{equation}
    s_j = \texttt{max}(|\textbf{X}_j|), \hspace{3mm}j = {1, 2, ..., C_i}
    \label{eq:scale-all}
\end{equation}

Where $C_i$ is the number of input channels in the input activation and $j$ corresponds to $j^{th}$ input channel. Note that since the range of activations varies for different input samples, the maximum value of each channel is estimated using $128$ calibration samples from the calibration dataset (see Section~\ref{sec:experiment-setup} for more details).
By dividing the input activation by the the scaling factor of Equation~\ref{eq:scale-all}, all channels of the scaled input activation would have the same range, making quantization of the scaled tensor to be very easy. However, this will migrate the difficulty of the quantization completely to the weight side of a linear layer. To address this issue, Xiao et al. proposed a scaling formula that balances the quantization difficulty of activations and weights:

\begin{equation}
    s_j = \texttt{max}(|\textbf{X}_j|)^\alpha/\texttt{max}(|\textbf{W}_j|)^{1-\alpha}, \hspace{3mm}j = {1, 2, ..., C_i}
    \label{eq:scale}
\end{equation}

Where $\alpha$ is a hyper-parameter that controls how much quantization difficulty we want to migrate from activations to weights. 
%For quantization to the MX format using SmoothQuant, we directly calculated the SmoothQuant scaling factors, skipping the additional calibration phase required for quantization to fixed-point formats. 
For more details on the SmoothQuant algorithm refer to Xiao et al.'s work~\cite{xiao2023smoothquant}.

\subsection{AWQ}
\label{sec:awq}
Activation-aware Weight Quantization (AWQ), is a weight-only quantization method for LLMs~\cite{lin2023awq}. In this algorithm, a small fraction (i.e., $0.1\%\texttt{-}1\%$) of salient weight channels are scaled up to reduce their relative quantization error:

\begin{equation}
    \textbf{Y} = \textbf{X}\textbf{W} \approx \textbf{X}\hat{\textbf{W}} \approx (\textbf{X}/s) (\hat{s\textbf{W})}
    \label{eq:awq}
\end{equation}

In Equation~\ref{eq:awq}, \textit{s} is a per-channel scaling factor for the salient weights. To determine the salient weights, AWQ refers to the activation distribution instead of the weight distribution, as weight channels corresponding to the outlier activations are more salient than other weights. The per-channel scaling factor is calculated using the following formula:

\begin{equation}
    s = s_\textbf{X}^\alpha, \hspace{3mm}\alpha \in [0, 1]
    \label{eq:awq-scale}
\end{equation}

Where $s_\textbf{X}$ is the average magnitude of activation (per-channel), and $\alpha$ is a hyper-parameter which balances the protection of salient and non-salient channels. 
% Similar to SmoothQuant, to make AWQ compatible with the MX format, we directly calculate the per-channel scaling factors, skipping the additional calibration phase required for fixed-point quantization. 
For more details on AWQ refer to Lin's et al. work~\cite{lin2023awq}

\begin{table}[t!] 
\centering
\footnotesize
\caption{Perplexity score on \emph{WikiText-2-test} for the Llama models, when quantized to MXINT formats with the block size of \emph{16} using different post-training quantization techniques. \textit{A}, \textit{W}, SQ, and RTN denote activation, weight, SmoothQuant, and round to nearest, respectively. \texttt{+}: GPTQ weight quantization is used.}
    \begin{tabular}{l|l|l||c|c|c}
    \toprule
    Bit-width  & Format      & Method & Llama2-7B      & Llama2-13B    & Llama3.1-8B      \\ \hline \hline
    A:16, W:16 & A:FP16, W:FP16  & N/A    & \textbf{5.12}  & \textbf{4.57} & \textbf{5.61}  \\ \hline
               & & RTN    & \textbf{5.12}  & 4.58          & \textbf{5.61}  \\
               & & GPTQ   & \textbf{5.12}  & 4.58          & \textbf{5.61}  \\
               & & SQ     & \textbf{5.12}  & \textbf{4.57} & \textbf{5.61}  \\
               & & AWQ    & \textbf{5.12}  & 4.58          & \textbf{5.61}  \\
    & \multirow{-5.5}{*}{A:MXINT8-16}  & SQ+  & \textbf{5.12}  & \textbf{4.57} & \textbf{5.61}  \\
    \multirow{-5.5}{*}{A:8, W:8} & \multirow{-4.5}{*}{W:MXINT8-16}  & AWQ+ & \textbf{5.12}  & \textbf{4.57}  & \textbf{5.61} \\ \hline
               & & RTN    & 5.40           & 4.72          & 6.22           \\ 
               & & GPTQ   & 5.41           & \textbf{4.68} & 5.97           \\
               & & SQ     & 5.33           & 4.74          & 6.19           \\
               & & AWQ    & 5.30           & 4.70          & 6.08           \\
    & \multirow{-5.5}{*}{A:MXINT8-16}  & SQ+  & 5.28        & 4.69          & 5.99          \\
    \multirow{-5.5}{*}{A:8, W:4} & \multirow{-4.5}{*}{W:MXINT4-16}  & AWQ+ & \textbf{5.27} & \textbf{4.68} & \textbf{5.95} \\ \hline
               & & RTN    & 6.58           & 5.52          & 10.16        \\ 
               & & GPTQ   & 6.84           & 5.09          & 7.29         \\
               & & SQ     & 6.56           & 5.49          & 10.54        \\
               & & AWQ    & 6.28           & 5.30          & 8.37         \\
    & \multirow{-5.5}{*}{A:MXINT8-16} & SQ+  & 5.90         & 5.14       & 7.35   \\
    \multirow{-5.5}{*}{A:8, W:3} & \multirow{-4.5}{*}{W:MXINT3-16} & AWQ+ & \textbf{5.84} & \textbf{5.09} & \textbf{7.14} \\ 
    
    \bottomrule
    \end{tabular}
\label{tab2:llama}
\end{table}

\subsection{Complete results of main result tables}
\label{sec:eval_harness_complete}
In Tables~\ref{tab:llama2-7b_llama2-13b_eval_harness},~\ref{tab:llama3-8b_eval_harness} and~\ref{tab:Qwen_eval_harness}, we present the complete set of downstream accuracy results for the two main Tables: ~\ref{tab:llama} and~\ref{tab:qwen2}. We evaluate the accuracy of the \emph{Llama2}, \emph{Llama3.1}, and \emph{Qwen2} model families on eight zero-shot commonsense reasoning tasks, including ARC-easy, ARC-challenge~\cite{clark2018arc}, BoolQ~\cite{clark2019boolq}, PIQA~\cite{bisk2020piqa}, SIQA~\cite{sap2019socialiqa}, HellaSwag~\cite{zellers2019hellaswag}, OBQA~\cite{obqa2018can}, and WinoGrande~\cite{sakaguchi2021winogrande}. These experiments investigate the effects of various quantization techniques—SmoothQuant, GPTQ, and AWQ—across different MXINT and fixed-point formats, providing a comprehensive view of how these techniques interact with each other and influence model performance under different quantization settings.

\begin{table*}[h!] 
\centering
\tiny
\caption{Accuracy on eight zero-shot common sense reasoning tasks, ARC-challenge, ARC-easy, BoolQ, HellaSwag, OBQA, PIQA, SIQA, and WinoGrande tasks, for \textit{Llama2-7B, and Llama2-13B}. A, W, SQ, and RTN denote activation, weight, SmoothQuant, and round to nearest, respectively. We used \textit{per-channel affine} quantization for the fixed-point formats. For the MXINT formats, we used block size of 128. \texttt{+}: GPTQ weight quantization is used. $\uparrow$: higher is better.}
\begin{tabular}{c|l|l||ccccccccc}
    \toprule
        \multirow{3}{*}{Bit-width} & \multirow{3}{*}{Format} & \multirow{3}{*}{Method} & \multicolumn{9}{c}{Llama2-7B}\\ \cdashline{4-12}
           &  &  & ARC-c & ARC-e & BoolQ & HellaS & OBQA & PIQA & SIQA & WinoG & Avg. \\
           &      &        & ($\uparrow$) & ($\uparrow$) & ($\uparrow$) & ($\uparrow$) & ($\uparrow$) & ($\uparrow$) & ($\uparrow$) & ($\uparrow$) & ($\uparrow$) \\ \hline 
         A:16,W:16  & A:FP16, W:FP16 & N/A  & \textbf{43.43} & \textbf{76.30} & \textbf{77.74} & \textbf{57.13} & \textbf{31.40} & \textbf{78.07} & \textbf{46.06} & \textbf{68.98} & \textbf{59.89} \\ \hline 
                       &             & RTN  & 43.69 & 76.22 & 77.52 & 57.07 & 31.20 & \textbf{78.07} & \textbf{46.42} & 68.82 & 59.88 \\
                       &             & GPTQ & 42.92 & 76.30 & 77.31 & 57.07 & 31.20 & 77.91 & 46.06 & 68.75 & 59.69 \\
                       &             & SQ   & 43.34 & 76.26 & 77.46 & 57.07 & 31.40 & 78.02 & 46.32 & 69.06 & 59.87 \\
                       &             & AWQ  & \textbf{44.03} & 76.26 & \textbf{77.71} & 57.17 & \textbf{31.80} & 77.80 & 45.91 & 69.06 & \textbf{59.97} \\
                       & \multirow{-5}{*}{A:MXINT8,} & SQ+ & 43.34 & 76.47 & 77.52 & \textbf{57.24} & 31.60 & 77.97 & 46.01 & 68.82 & 59.87  \\ 
                       & \multirow{-5}{*}{W:MXINT8} & AWQ+ & 43.52 & \textbf{76.52} & 77.46 & 57.13 & 31.60 & 78.02 & 45.85 & \textbf{69.14} & 59.90 \\ \cdashline{2-12}
                       &            & RTN  & \textbf{43.69} & 76.52 & 67.98 & 56.74 & 31.80 & \textbf{77.69} & 46.06 & 67.88 & 58.54 \\ 
                       &            & GPTQ & 42.58 & 76.47 & \textbf{68.47} & \textbf{56.78} & 31.40 & 77.26 & 45.91 & \textbf{69.30} & 58.52 \\
                       &            & SQ   & 42.92 & \textbf{76.77} & 68.35 & 56.65 & \textbf{32.60} & 77.48 & 46.21 & 68.51 & 58.69 \\
                       &            & AWQ  & 42.49 & 76.43 & 67.37 & 56.32 & 31.80 & 76.71 & 45.70 & 68.98 & 58.23 \\
    \multirow{-11}{*}{A:8,} & \multirow{-5}{*}{A:INT8,} & SQ+ &  43.17 & 76.52 & 68.41 & 56.68 & 32.00 & 77.53 & \textbf{46.42} & 68.98 & \textbf{58.71} \\
    \multirow{-11}{*}{W:8}  & \multirow{-5}{*}{W:INT8}  & AWQ+ & 42.24 & 75.88 & 67.13 & 56.41 & 31.80 & 77.04 & 46.01 & 68.67 & 58.15 \\ \hline
                       &            & RTN  & \textbf{42.66} & \textbf{74.87} & \textbf{75.47} & \textbf{56.08} & \textbf{32.40} & 76.93 & 44.52 & \textbf{68.98} & \textbf{58.99} \\ 
                       &            & GPTQ & 40.27 & 72.56 & 74.71 & 54.86 & 31.00 & 76.33 & 44.22 & 67.32 & 57.66 \\
                       &            & SQ   & 40.70 & 74.03 & 71.93 & 54.63 & 30.20 & 77.15 & 44.17 & 67.88 & 57.59 \\
                       &            & AWQ  & 41.55 & 73.95 & 75.11 & 55.90 & 31.60 & 76.99 & 44.63 & 67.72 & 58.43 \\
                       & \multirow{-5}{*}{A:MXINT8,} & SQ+ & 41.72 & 73.61 & 73.12 & 54.55 & 30.40 & \textbf{77.58} & 44.22 & 67.80 & 57.88  \\
                       & \multirow{-5}{*}{W:MXINT4} & AWQ+ & 40.70 & 73.57 & 74.31 & 55.89 & 30.80 & 77.48 & \textbf{44.68} & 68.27 & 58.21 \\ \cdashline{2-12}
                       &            & RTN  & 41.38 & 73.15 & 66.45 & \textbf{55.29} & 29.80 & 76.88 & 43.71 & 66.85 & 56.69 \\ 
                       &            & GPTQ & 40.96 & 73.19 & 66.27 & 54.33 & 28.80 & 76.61 & \textbf{44.27} & 67.80 & 56.53 \\
                       &            & SQ   & 39.16 & 73.06 & 68.65 & 52.09 & 26.80 & 74.43 & 42.53 & 65.43 & 55.27 \\
                       &            & AWQ  & \textbf{41.72} & 74.83 & 68.26 & 54.98 & \textbf{31.20} & \textbf{77.20} & 44.06 & \textbf{68.82} & \textbf{57.64} \\
    \multirow{-11}{*}{A:8,} & \multirow{-5}{*}{A:INT8,} & SQ+ & 39.51 & 71.63 & \textbf{71.96} & 53.33 & 29.60 & 75.63 & 43.71 & 66.46 & 56.48 \\
    \multirow{-11}{*}{W:4}  & \multirow{-5}{*}{W:INT4} & AWQ+ & 40.36 & \textbf{74.87} & 68.29 & 54.21 & 30.20 & 76.77 & 44.11 & 68.43 & 57.16 \\ \hline
                       &            & RTN  & 33.53 & 65.07 & 60.76 & 48.92 & 27.80 & 73.18 & 40.79 & 62.51 & 51.57 \\ 
                       &            & GPTQ & 30.29 & 63.47 & 66.06 & 44.02 & 21.00 & 70.57 & 40.17 & 59.51 & 49.38 \\
                       &            & SQ   & 28.50 & 55.60 & 56.30 & 41.30 & 21.20 & 67.68 & 39.00 & 57.70 & 45.91 \\
                       &            & AWQ  & \textbf{34.81} & \textbf{68.06} & \textbf{66.39} & \textbf{50.68} & \textbf{28.20} & \textbf{75.73} & \textbf{42.89} & \textbf{64.64} & \textbf{53.92} \\
                       & \multirow{-5}{*}{A:MXINT8,} & SQ+ & 31.06 & 59.76 & 65.05 & 44.62 & 23.00 & 69.26 & 39.56 & 61.40 & 49.21 \\
                       & \multirow{-5}{*}{W:MXINT3} & AWQ+ & 31.40 & 64.94 & 65.66 & 46.33 & 24.40 & 72.31 & 41.20 & 62.04 & 51.03 \\ \cdashline{2-12}
                       &            & RTN  & 21.25 & 25.93 & 38.32 & 25.78 & 15.40 & 51.63 & 34.24 & 47.91 & 32.56 \\ 
                       &            & GPTQ & 18.86 & 34.85 & 57.52 & 29.94 & 14.60 & 58.92 & 35.06 & 53.04 & 37.85 \\
                       &            & SQ   & - & - & - & - & - & - & - & - & -  \\
                       &            & AWQ  & 20.73 & 31.99 & 57.92 & 29.07 & 15.60 & 55.60 & 34.80 & 51.46 & 37.15 \\
    \multirow{-11}{*}{A:8,} & \multirow{-5}{*}{A:INT8,} & SQ+ & - & - & - & - & - & - & - & - & - \\
    \multirow{-11}{*}{W:3}  & \multirow{-5}{*}{W:INT3} & AWQ+ & \textbf{23.46} & \textbf{52.78} & \textbf{62.42} & \textbf{36.72} & \textbf{16.20} & \textbf{65.02} & \textbf{38.59} & \textbf{56.99} & \textbf{44.02} \\
    \hline \hline
    %%%% llama2-13%%%%
    \multirow{3}{*}{Bit-width} & \multirow{3}{*}{Format} & \multirow{3}{*}{Method} & \multicolumn{9}{c}{Llama2-13B}\\ \cdashline{4-12}
           &  &  & ARC-c & ARC-e & BoolQ & HellaS & OBQA & PIQA & SIQA & WinoG & Avg. \\
           &      &        & ($\uparrow$) & ($\uparrow$) & ($\uparrow$) & ($\uparrow$) & ($\uparrow$) & ($\uparrow$) & ($\uparrow$) & ($\uparrow$) & ($\uparrow$) \\ \hline 
         A:16,W:16  & A:FP16, W:FP16 & N/A  & \textbf{48.46} & \textbf{79.42} & \textbf{80.58} & \textbf{60.04} & \textbf{35.20} & \textbf{79.11} & \textbf{47.29} & \textbf{72.30} & \textbf{62.80} \\ \hline 
                       &             & RTN  & 47.87 & 79.34 & 80.55 & 60.08 & 34.40 & 79.00 & 47.08 & 71.82 & 62.52 \\
                       &             & GPTQ & \textbf{48.12} & 79.50 & 80.37 & 60.03 & 34.40 & \textbf{79.38} & 47.08 & 72.06 & 62.62 \\
                       &             & SQ   & 47.95 & \textbf{79.55} & 80.67 & 60.01 & \textbf{35.20} & 79.16 & 47.34 & 71.90 & 62.72 \\
                       &             & AWQ  & 47.95 & \textbf{79.55} & \textbf{80.70} & \textbf{60.18} & 35.00 & 79.22 & \textbf{47.49} & \textbf{72.30} & \textbf{62.80} \\
                       & \multirow{-5}{*}{A:MXINT8,} & SQ+ & - & - & - & - & - & - & - & - & - \\ 
                       & \multirow{-5}{*}{W:MXINT8} & AWQ+ & 47.61 & 79.46 & 80.64 & 60.13 & 34.60 & 79.33 & 47.44 & 71.90 & 62.64 \\ \cdashline{2-12}
                       &            & RTN  & 48.29 & 78.79 & 73.06 & 59.69 & 34.40 & 78.29 & 46.83 & \textbf{72.69} & 61.50 \\ 
                       &            & GPTQ & \textbf{48.89} & 79.25 & 73.06 & \textbf{59.75} & 34.40 & 78.29 & 46.52 & \textbf{72.69} & \textbf{61.61} \\
                       &            & SQ   & 48.46 & 78.91 & 72.97 & 59.72 & 34.60 & \textbf{78.51} & \textbf{46.98} & 72.30 & 61.56 \\
                       &            & AWQ  & 47.87 & \textbf{79.80} & \textbf{73.30} & 59.68 & \textbf{34.80} & \textbf{78.51} & 46.16 & 72.53 & 61.58 \\
    \multirow{-11}{*}{A:8,} & \multirow{-5}{*}{A:INT8,} & SQ+ & - & - & - & - & - & - & - & - & -  \\
    \multirow{-11}{*}{W:8}  & \multirow{-5}{*}{W:INT8}  & AWQ+ & 47.70 & 79.21 & 73.27 & 59.61 & 34.60 & 78.45 & 46.57 & 72.30 & 61.46 \\ \hline
                       &            & RTN  & 46.50 & 78.58 & \textbf{80.76} & 58.78 & 34.00 & 78.24 & 46.26 & 71.51 & 61.83 \\ 
                       &            & GPTQ & 46.33 & \textbf{78.70} & 77.09 & 58.59 & \textbf{35.20} & 77.91 & 46.62 & 71.82 & 61.53 \\
                       &            & SQ   & 44.54 & 77.10 & 78.47 & 58.16 & 34.60 & 77.86 & 45.50 & 72.22 & 61.06 \\
                       &            & AWQ  & \textbf{47.61} & 78.24 & 80.34 & 58.99 & 34.00 & \textbf{78.84} & 46.42 & \textbf{72.45} & \textbf{62.11} \\
                       & \multirow{-5}{*}{A:MXINT8,} & SQ+ & - & - & - & - & - & - & - & - & -  \\
                       & \multirow{-5}{*}{W:MXINT4} & AWQ+ & 46.67 & 78.37 & 79.88 & \textbf{59.01} & 33.00 & 78.56 & \textbf{47.19} & 71.03 & 61.71 \\ \cdashline{2-12}
                       &            & RTN  & 45.31 & \textbf{79.12} & 71.01 & 58.05 & 33.20 & \textbf{78.02} & 45.45 & 70.96 & 60.14 \\ 
                       &            & GPTQ & 43.52 & 77.61 & 70.18 & 57.98 & 33.80 & 77.75 & 45.09 & 70.40 & 59.54 \\
                       &            & SQ   & 41.30 & 75.08 & 73.12 & 54.52 & 28.20 & 75.46 & 42.73 & 68.27 & 57.34 \\
                       &            & AWQ  & \textbf{46.93} & 79.04 & \textbf{73.61} & 58.15 & \textbf{34.40} & 77.97 & \textbf{45.80} & 70.96 & \textbf{60.86} \\
    \multirow{-11}{*}{A:8,} & \multirow{-5}{*}{A:INT8,} & SQ+ & - & - & - & - & - & - & - & - & - \\
    \multirow{-11}{*}{W:4}  & \multirow{-5}{*}{W:INT4} & AWQ+ & 46.42 & 78.37 & 69.66 & \textbf{58.43} & 33.80 & 77.58 & 45.34 & \textbf{71.35} & 60.12 \\ \hline
                       &            & RTN  & 38.82 & 70.58 & 63.06 & 52.76 & 28.60 & 74.43 & 41.61 & 68.03 & 54.74 \\ 
                       &            & GPTQ & 38.14 & 71.51 & 64.07 & 52.43 & \textbf{30.20} & 75.35 & \textbf{43.40} & 69.06 & 55.52 \\
                       &            & SQ   & 31.83 & 64.06 & 66.97 & 45.32 & 21.20 & 70.62 & 38.64 & 61.40 & 50.00 \\
                       &            & AWQ  & \textbf{40.78} & \textbf{72.81} & 72.17 & \textbf{54.40} & 28.40 & 75.57 & 41.81 & \textbf{69.38} & \textbf{56.92} \\
                       & \multirow{-5}{*}{A:MXINT8,} & SQ+ & - & - & - & - & - & - & - & - & - \\
                       & \multirow{-5}{*}{W:MXINT3} & AWQ+ & 37.29 & 71.68 & \textbf{72.87} & 52.67 & 28.20 & \textbf{76.33} & 41.20 & 66.46 & 55.84  \\ \cdashline{2-12}
                       &            & RTN  & 19.45 & 33.38 & 56.18 & 29.15 & 13.20 & 55.11 & 35.11 & 50.83 & 36.55 \\ 
                       &            & GPTQ & 26.45 & 59.01 & 63.64 & 38.94 & 20.40 & 64.80 & 38.33 & 57.22 & 46.10 \\
                       &            & SQ   & 23.12 & 25.51 & 39.45 & 25.65 & 13.40 & 52.45 & 35.11 & 48.62 & 32.91 \\
                       &            & AWQ  & \textbf{33.70} & 66.79 & 63.06 & 44.65 & 25.20 & 70.13 & 41.10 & 62.04 & 50.83 \\
    \multirow{-11}{*}{A:8,} & \multirow{-5}{*}{A:INT8,} & SQ+ & - & - & - & - & - & - & - & - & - \\
    \multirow{-11}{*}{W:3}  & \multirow{-5}{*}{W:INT3} & AWQ+ & 33.11 & \textbf{67.55} & \textbf{64.16} & \textbf{46.88} & \textbf{25.60} & \textbf{71.00} & \textbf{41.61} & \textbf{64.33} & \textbf{51.78} \\
    \bottomrule

\end{tabular}
\label{tab:llama2-7b_llama2-13b_eval_harness}
\end{table*}

\begin{table*}[h!] 
\centering
\scriptsize
\caption{Accuracy on eight zero-shot common sense reasoning tasks, ARC-challenge, ARC-easy, BoolQ, HellaSwag, OBQA, PIQA, SIQA, and WinoGrande tasks, for \textit{Llama3.1-8B}. A, W, SQ, and RTN denote activation, weight, SmoothQuant, and round to nearest, respectively. We used \textit{per-channel affine} quantization for the fixed-point formats. For the MXINT formats, we used block size of 128. \texttt{+}: GPTQ weight quantization is used. $\uparrow$: higher is better.}
\begin{tabular}{c|l|l||ccccccccc}
    \toprule
          Bit & \multirow{2}{*}{Format} & \multirow{2}{*}{Method} & ARC-c & ARC-e & BoolQ & HellaS & OBQA & PIQA & SIQA & WinoG & Avg. \\
        Width  &      &        & ($\uparrow$) & ($\uparrow$) & ($\uparrow$) & ($\uparrow$) & ($\uparrow$) & ($\uparrow$) & ($\uparrow$) & ($\uparrow$) & ($\uparrow$) \\ \hline \hline
          A:16,  & A:FP16, & \multirow{2}{*}{N/A}  & \multirow{2}{*}{\textbf{51.19}} & \multirow{2}{*}{\textbf{81.44}} & \multirow{2}{*}{\textbf{82.05}} & \multirow{2}{*}{\textbf{60.01}} & \multirow{2}{*}{\textbf{33.40}} & \multirow{2}{*}{\textbf{80.03}} & \multirow{2}{*}{\textbf{47.13}} & \multirow{2}{*}{\textbf{73.56}} & \multirow{2}{*}{63.60\textbf{}} \\
          W:16   & W:FP16  &  &  \\\hline
                       &             & RTN  & 50.94 & \textbf{81.78} & \textbf{82.48} & 60.03 & 33.40 & 79.82 & 46.78 & 73.48 & 63.59 \\
                       &             & GPTQ & 50.94 & 81.73 & 82.29 & 60.02 & 33.00 & 79.82 & 47.08 & \textbf{73.95} & 63.60 \\
                       &             & SQ   & 51.11 & 81.36 & 82.14 & \textbf{60.09} & 33.60 & \textbf{80.14} & 46.83 & 73.32 & 63.57 \\
                       &             & AWQ  & 51.11 & 81.44 & 82.45 & 60.02 & 33.40 & 80.09 & \textbf{47.29} & 73.48 & \textbf{63.66} \\
                       & \multirow{-5}{*}{A:MXINT8,} & SQ+ & 50.94 & 81.36 & 82.11 & 60.05 & \textbf{33.80} & 79.71 & 46.72 & 73.88 & 63.57  \\ 
                       & \multirow{-5}{*}{W:MXINT8} & AWQ+ & \textbf{51.37} & 81.23 & 82.45 & 60.01 & 33.20 & 79.98 & 47.08 & 73.16 & 63.56 \\ \cdashline{2-12}
                       &            & RTN  & 50.26 & 81.94 & 79.08 & 59.43 & 34.20 & 79.54 & 47.03 & 72.30 & 62.97 \\ 
                       &            & GPTQ & 50.17 & 81.52 & \textbf{79.30} & \textbf{59.54} & 34.40 & 79.54 & 46.88 & 72.61 & 63.00 \\
                       &            & SQ   & \textbf{50.51} & \textbf{81.99} & 79.17 & 59.49 & 33.80 & 79.43 & \textbf{47.49} & 74.11 & \textbf{63.25} \\
                       &            & AWQ  & 50.09 & 81.52 & 75.35 & 59.44 & 34.40 & 80.09 & 46.52 & \textbf{74.35} & 62.72 \\
    \multirow{-11}{*}{A:8,} & \multirow{-5}{*}{A:INT8,} & SQ+ & 50.43 & 81.40 & 78.93 & 59.45 & \textbf{35.00} & 79.76 & 47.13 & 73.01 & 63.14 \\
    \multirow{-11}{*}{W:8}  & \multirow{-5}{*}{W:INT8}  & AWQ+ & 49.15 & 81.90 & 74.83 & 59.37 & 34.20 & \textbf{80.41} & 46.93 & 73.40 & 62.52 \\ \hline
                       &            & RTN  & 44.80 & 76.52 & \textbf{80.61} & 57.63 & \textbf{32.20} & 77.91 & 45.85 & \textbf{72.22} & 60.97 \\ 
                       &            & GPTQ & 45.82 & 77.53 & 78.50 & 57.28 & 31.60 & 78.40 & 46.26 & 71.11 & 60.81 \\
                       &            & SQ   & 46.42 & 77.74 & 74.25 & 55.55 & 28.80 & 77.69 & 45.60 & 70.09 & 59.52 \\
                       &            & AWQ  & \textbf{47.27} & \textbf{79.55} & 78.69 & \textbf{58.22} & 31.60 & \textbf{79.05} & 46.78 & 71.03 & \textbf{61.52} \\
                       & \multirow{-5}{*}{A:MXINT8,} & SQ+ &  45.48 & 77.31 & 75.75 & 56.56 & 31.80 & 77.97 & 46.52 & 69.30 & 60.09 \\
                       & \multirow{-5}{*}{W:MXINT4} & AWQ+ & 46.93 & 77.90 & 78.53 & 57.95 & 30.80 & 78.67 & \textbf{46.88} & 71.74 & 61.18 \\ \cdashline{2-12}
                       &            & RTN  & 45.22 & 75.67 & 76.97 & 54.89 & 31.40 & 77.86 & 45.34 & 70.48 & 59.73 \\ 
                       &            & GPTQ & 42.49 & 74.41 & 73.21 & 52.29 & 29.40 & 71.65 & 41.10 & 67.17 & 56.46 \\
                       &            & SQ   & 38.31 & 70.58 & 66.21 & 50.80 & 26.00 & 73.72 & 43.50 & 66.93 & 54.51 \\
                       &            & AWQ  & \textbf{47.87} & \textbf{79.34} & \textbf{77.98} & \textbf{57.70} & \textbf{34.60} & \textbf{78.89} & \textbf{47.39} & \textbf{71.35} & \textbf{61.89}   \\
    \multirow{-11}{*}{A:8,} & \multirow{-5}{*}{A:INT8,} & SQ+ & 40.87 & 72.10 & 68.53 & 51.52 & 29.00 & 75.46 & 44.11 & 68.82 & 56.30 \\
    \multirow{-11}{*}{W:4}  & \multirow{-5}{*}{W:INT4} & AWQ+ & 45.56 & 79.04 & 76.67 & 57.66 & 31.80 & 78.35 & 46.32 & 71.27 & 60.83 \\ \hline
                       &            & RTN  & 24.23 & 45.12 & 54.92 & 35.12 & 18.20 & 62.51 & 37.92 & 54.14 & 41.52 \\ 
                       &            & GPTQ & 25.09 & 48.36 & 57.19 & 39.67 & 18.40 & 64.58 & 38.13 & 57.06 & 43.56 \\
                       &            & SQ   & 19.20 & 27.95 & 39.30 & 26.86 & 13.40 & 54.79 & 33.67 & 50.12 & 33.16 \\
                       &            & AWQ  & \textbf{34.22} & \textbf{64.10} & 58.72 & 45.95 & \textbf{23.80} & \textbf{71.71} & \textbf{42.17} & 61.25 & \textbf{50.24} \\
                       & \multirow{-5}{*}{A:MXINT8,} & SQ+ & 20.48 & 41.50 & \textbf{60.09} & 36.91 & 16.20 & 60.72 & 37.92 & 53.67 & 40.94 \\
                       & \multirow{-5}{*}{W:MXINT3} & AWQ+ & 31.48 & 61.36 & 57.77 & \textbf{47.17} & 22.60 & 69.37 & 40.84 & \textbf{62.43} & 49.13 \\ \cdashline{2-12}
                       &            & RTN  & 19.62 & 27.02 & 38.72 & 26.09 & 12.00 & 54.95 & 34.49 & 48.70 & 32.70 \\ 
                       &            & GPTQ & 18.60 & 26.94 & 38.90 & 26.75 & 13.40 & 53.92 & 33.06 & 51.22 & 32.85 \\
                       &            & SQ   & \textbf{22.27} & 25.34 & 46.18 & 25.58 & 13.20 & 53.10 & 35.36 & 52.57 & 34.20 \\
                       &            & AWQ  & 20.39 & 40.15 & \textbf{60.98} & 28.93 & \textbf{16.20} & 60.01 & 35.93 & 50.83 & 39.18 \\
    \multirow{-11}{*}{A:8,} & \multirow{-5}{*}{A:INT8,} & SQ+ & 19.97 & 26.60 & 38.93 & 26.28 & 15.20 & 52.12 & 33.52 & 48.54 & 32.64 \\
    \multirow{-11}{*}{W:3}  & \multirow{-5}{*}{W:INT3} & AWQ+ & 21.67 & \textbf{47.18} & 60.80 & \textbf{32.18} & \textbf{16.20} & \textbf{61.75} & \textbf{37.51} & \textbf{55.17} & \textbf{41.56} \\
    \bottomrule

\end{tabular}
\label{tab:llama3-8b_eval_harness}
\end{table*}

\begin{table*}[h!] 
\centering
\tiny
\caption{Accuracy on eight zero-shot common sense reasoning tasks, ARC-challenge, ARC-easy, BoolQ, HellaSwag, OBQA, PIQA, SIQA, and WinoGrande tasks, for \textit{Qwen2-1.5B, and Qwen2-7B}. A, W, SQ, and RTN denote activation, weight, SmoothQuant, and round to nearest, respectively. We used \textit{per-channel affine} quantization for the fixed-point formats. For the MXINT formats, we used block size of 128. \texttt{+}: GPTQ weight quantization is used. $\uparrow$: higher is better.}
\begin{tabular}{c|l|l||ccccccccc}
    \toprule
        \multirow{3}{*}{Bit-width} & \multirow{3}{*}{Format} & \multirow{3}{*}{Method} & \multicolumn{9}{c}{Qwen2-1.5B}\\ \cdashline{4-12}
           &  &  & ARC-c & ARC-e & BoolQ & HellaS & OBQA & PIQA & SIQA & WinoG & Avg. \\
           &      &        & ($\uparrow$) & ($\uparrow$) & ($\uparrow$) & ($\uparrow$) & ($\uparrow$) & ($\uparrow$) & ($\uparrow$) & ($\uparrow$) & ($\uparrow$) \\ \hline 
         A:16,W:16  & A:FP16, W:FP16 & N/A  & \textbf{33.62} & \textbf{66.12} & \textbf{72.78} & \textbf{48.64} & \textbf{27.00} & \textbf{75.30} & \textbf{45.80} & \textbf{65.90} & \textbf{54.40} \\ \hline 
                       &             & RTN  & 33.53 & 66.33 & \textbf{72.39} & 48.51 & 27.00 & 75.68 & 45.96 & 66.06 & \textbf{54.43} \\
                       &             & GPTQ & \textbf{33.62} & 66.20 & 72.29 & 48.51 & 26.20 & 75.63 & \textbf{46.16} & 65.98 & 54.32 \\
                       &             & SQ   & 33.19 & 66.29 & 72.29 & 48.63 & 26.00 & 75.68 & 45.75 & \textbf{66.69} & 54.32 \\
                       &             & AWQ  & 33.45 & \textbf{66.79} & 72.02 & 48.64 & 26.20 & 75.57 & 45.85 & 65.27 & 54.22 \\
                       & \multirow{-5}{*}{A:MXINT8,} & SQ+ & 33.36 & 66.04 & 71.74 & \textbf{48.71} & 26.60 & \textbf{75.73} & 45.55 & 65.75 & 54.18 \\ 
                       & \multirow{-5}{*}{W:MXINT8} & AWQ+ & 32.94 & 66.25 & 72.29 & 48.55 & \textbf{27.20} & 75.14 & 46.11 & 65.90 & 54.30 \\ \cdashline{2-12}
                       &            & RTN  & 33.79 & \textbf{67.72} & \textbf{72.29} & 48.02 & 27.00 & 74.54 & 46.83 & 64.40 & 54.32 \\ 
                       &            & GPTQ & 33.53 & 66.88 & 71.90 & \textbf{48.19} & 27.40 & \textbf{75.24} & 46.21 & 65.59 & 54.37 \\
                       &            & SQ   & 33.36 & 66.41 & 72.23 & 48.02 & 26.00 & 74.97 & 46.37 & 65.51 & 54.11 \\
                       &            & AWQ  & 33.70 & 67.38 & 67.00 & 47.75 & 27.40 & 74.76 & 46.67 & 64.88 & 53.69 \\
    \multirow{-11}{*}{A:8,} & \multirow{-5}{*}{A:INT8,} & SQ+ & \textbf{33.87} & 66.96 & 72.05 & 48.05 & 26.60 & 74.37 & \textbf{46.93} & \textbf{66.85} & \textbf{54.46} \\
    \multirow{-11}{*}{W:8}  & \multirow{-5}{*}{W:INT8}  & AWQ+ & 33.11 & 67.34 & 66.97 & 47.85 & \textbf{27.60} & 74.86 & 46.72 & 64.48 & 53.62 \\ \hline
                       &            & RTN  & 32.00 & 65.07 & 69.17 & 46.29 & 26.20 & 73.01 & \textbf{46.01} & 61.09 & 52.35 \\ 
                       &            & GPTQ & \textbf{33.96} & \textbf{67.68} & 65.84 & 46.71 & 24.80 & 73.99 & 45.24 & \textbf{63.46} & 52.71 \\
                       &            & SQ   & 30.03 & 59.13 & 53.24 & 45.52 & \textbf{26.80} & 73.61 & 44.42 & 61.88 & 49.33 \\
                       &            & AWQ  & 31.66 & 62.25 & 70.80 & 46.88 & 25.60 & 73.67 & 44.93 & 62.04 & 52.23 \\
                       & \multirow{-5}{*}{A:MXINT8,} & SQ+ & 30.29 & 60.48 & 67.92 & 45.38 & 24.00 & 72.91 & 42.73 & 62.12 & 50.73 \\
                       & \multirow{-5}{*}{W:MXINT4} & AWQ+ & 31.66 & 65.57 & \textbf{72.14} & \textbf{46.90} & 24.80 & \textbf{74.05} & 45.34 & 62.75 & \textbf{52.90} \\ \cdashline{2-12}
                       &            & RTN  & 30.89 & \textbf{65.19} & 67.92 & 45.27 & 25.00 & 72.47 & 42.73 & 62.67 & 51.52 \\ 
                       &            & GPTQ & 32.68 & 63.51 & 67.80 & 45.74 & 24.20 & 71.82 & 43.81 & 61.48 & 51.38 \\
                       &            & SQ   & 28.07 & 59.97 & 66.79 & 43.22 & 21.80 & 70.46 & 41.04 & 59.75 & 48.89 \\
                       &            & AWQ  & \textbf{33.02} & 64.27 & 65.20 & 45.71 & \textbf{26.00} & \textbf{73.83} & \textbf{45.75} & 62.43 & \textbf{52.03} \\
    \multirow{-11}{*}{A:8,} & \multirow{-5}{*}{A:INT8,} & SQ+ & 30.89 & 63.26 & 67.58 & 43.77 & 24.80 & 71.55 & 43.35 & 61.33 & 50.81 \\
    \multirow{-11}{*}{W:4}  & \multirow{-5}{*}{W:INT4} & AWQ+ & 31.40 & 59.18 & \textbf{69.30} & \textbf{45.93} & 24.60 & 72.69 & 43.14 & \textbf{65.90} & 51.52 \\ \hline
                       &            & RTN  & 24.66 & 40.78 & 46.64 & 36.63 & 20.20 & 64.58 & \textbf{38.79} & 54.70 & 40.87 \\ 
                       &            & GPTQ & 25.43 & 47.22 & 50.64 & 37.95 & 20.20 & 64.91 & 37.92 & \textbf{56.51} & 42.60 \\
                       &            & SQ   & 21.08 & 36.70 & 54.40 & 30.08 & 15.20 & 55.88 & 34.14 & 49.41 & 37.11 \\
                       &            & AWQ  & \textbf{27.22} & \textbf{55.30} & \textbf{63.30} & \textbf{40.42} & 17.80 & \textbf{67.95} & 38.33 & 56.20 & \textbf{45.82} \\
                       & \multirow{-5}{*}{A:MXINT8,} & SQ+ & 20.65 & 35.77 & 51.53 & 33.36 & 14.60 & 58.87 & 36.44 & 52.64 & 37.98 \\
                       & \multirow{-5}{*}{W:MXINT3} & AWQ+ & 25.94 & 45.50 & 55.66 & 40.25 & \textbf{20.40} & \textbf{67.95} & 37.00 & 56.12 & 43.60 \\ \cdashline{2-12}
                       &            & RTN  & 20.14 & 27.65 & 49.14 & 26.50 & 16.00 & 51.31 & 34.03 & 51.46 & 34.53 \\ 
                       &            & GPTQ & 21.84 & 35.23 & 44.92 & 32.16 & 15.80 & 60.12 & 35.77 & 51.46 & 37.16 \\
                       &            & SQ   & 21.59 & 26.26 & 43.27 & 26.12 & 13.20 & 52.56 & 33.42 & 47.12 & 32.94 \\
                       &            & AWQ  & \textbf{25.60} & \textbf{49.49} & \textbf{62.69} & 37.69 & \textbf{18.40} & 63.76 & 38.59 & 54.70 & \textbf{43.87} \\
    \multirow{-11}{*}{A:8,} & \multirow{-5}{*}{A:INT8,} & SQ+ & 21.08 & 25.55 & 44.62 & 26.26 & 13.00 & 52.29 & 33.93 & 50.20 & 33.36 \\
    \multirow{-11}{*}{W:3}  & \multirow{-5}{*}{W:INT3} & AWQ+ & 23.46 & 46.04 & 59.17 & \textbf{38.90} & 17.20 & \textbf{66.10} & \textbf{40.02} & \textbf{58.72} & 43.70 \\
    \hline \hline
    %%%% Qwen2-7B%%%%
    \multirow{3}{*}{Bit-width} & \multirow{3}{*}{Format} & \multirow{3}{*}{Method} & \multicolumn{9}{c}{Qwen2-7B}\\ \cdashline{4-12}
           &  &  & ARC-c & ARC-e & BoolQ & HellaS & OBQA & PIQA & SIQA & WinoG & Avg. \\
           &      &        & ($\uparrow$) & ($\uparrow$) & ($\uparrow$) & ($\uparrow$) & ($\uparrow$) & ($\uparrow$) & ($\uparrow$) & ($\uparrow$) & ($\uparrow$) \\ \hline 
         A:16,W:16  & A:FP16, W:FP16 & N/A  & \textbf{48.81} & \textbf{79.21} & \textbf{84.83} & \textbf{59.32} & \textbf{34.80} & \textbf{79.82} & \textbf{48.52} & \textbf{72.30} & \textbf{63.45} \\ \hline 
                       &             & RTN  & 48.38 & 79.12 & 84.65 & 59.18 & \textbf{34.60} & 79.60 & 48.62 & 72.06 & 63.28 \\
                       &             & GPTQ & 48.21 & \textbf{79.63} & \textbf{85.05} & 59.12 & 34.00 & 79.60 & 48.36 & 71.43 & 63.17 \\
                       &             & SQ   & \textbf{49.15} & 79.25 & 84.83 & 59.07 & 34.20 & 79.65 & \textbf{48.67} & 71.82 & 63.33 \\
                       &             & AWQ  & 48.38 & 79.34 & 84.68 & \textbf{59.32} & 34.20 & \textbf{79.92} & 48.62 & 71.51 & 63.25 \\
                       & \multirow{-5}{*}{A:MXINT8,} & SQ+ & 48.12 & 79.21 & 84.92 & 59.12 & 34.00 & 79.71 & 48.46 & \textbf{72.14} & 63.21 \\ 
                       & \multirow{-5}{*}{W:MXINT8} & AWQ+ & 48.38 & \textbf{79.63} & 84.77 & 59.25 & 34.20 & \textbf{79.92} & \textbf{48.67} & 71.98 & \textbf{63.35} \\ \cdashline{2-12}
                       &            & RTN  & 48.04 & 79.67 & \textbf{85.08} & \textbf{58.00} & 33.20 & 78.89 & 47.39 & 72.06 & 62.79 \\ 
                       &            & GPTQ & \textbf{48.89} & 79.34 & 84.50 & 57.90 & 34.00 & 79.00 & 48.06 & 72.22 & 62.99 \\
                       &            & SQ   & 48.63 & \textbf{80.22} & \textbf{85.08} & 57.91 & 33.60 & \textbf{79.33} & \textbf{49.03} & 70.56 & 63.04 \\
                       &            & AWQ  & 48.55 & 79.55 & 83.88 & 57.42 & 33.40 & 78.78 & 48.36 & 71.43 & 62.67 \\
    \multirow{-11}{*}{A:8,} & \multirow{-5}{*}{A:INT8,} & SQ+ & 48.55 & 79.21 & 84.65 & 57.86 & \textbf{34.20} & 79.00 & 48.16 & \textbf{72.77} & \textbf{63.05} \\
    \multirow{-11}{*}{W:8}  & \multirow{-5}{*}{W:INT8}  & AWQ+ & \textbf{48.89} & 79.38 & 83.91 & 57.75 & 33.20 & 78.73 & 48.72 & 72.45 & 62.88 \\ \hline
                       &            & RTN  & 43.86 & 74.20 & 82.20 & 51.83 & 31.60 & 75.63 & 48.93 & 65.98 & 59.28 \\ 
                       &            & GPTQ & \textbf{48.72} & \textbf{80.26} & 82.60 & 57.27 & 30.40 & 78.56 & 48.36 & 71.74 & \textbf{62.24} \\
                       &            & SQ   & 47.61 & 77.86 & 82.05 & 56.23 & \textbf{32.00} & 78.67 & 47.85 & 69.06 & 61.42 \\
                       &            & AWQ  & 46.50 & 76.01 & 82.97 & \textbf{58.00} & \textbf{32.00} & 78.78 & 48.62 & 69.93 & 61.60 \\
                       & \multirow{-5}{*}{A:MXINT8,} & SQ+ & 44.54 & 77.82 & 82.66 & 56.45 & 31.60 & 77.75 & 48.31 & 70.40 & 61.19 \\
                       & \multirow{-5}{*}{W:MXINT4} & AWQ+ & 47.01 & 78.11 & \textbf{83.52} & \textbf{58.00} & 30.80 & \textbf{78.84} & \textbf{48.98} & \textbf{71.82} & 62.13 \\ \cdashline{2-12}
                       &            & RTN  & 37.80 & 65.45 & 72.60 & 36.48 & 30.80 & 69.04 & 45.04 & 56.43 & 51.70 \\ 
                       &            & GPTQ & 44.20 & 76.43 & \textbf{82.35} & 55.44 & \textbf{31.60} & 77.48 & 48.21 & 71.35 & 60.88 \\
                       &            & SQ   & 39.68 & 66.75 & 73.33 & 44.00 & 27.60 & 72.25 & 43.14 & 63.77 & 53.82 \\
                       &            & AWQ  & \textbf{47.87} & \textbf{78.16} & 78.53 & 55.58 & 34.00 & \textbf{78.35} & \textbf{49.64} & 71.43 & \textbf{61.69} \\
    \multirow{-11}{*}{A:8,} & \multirow{-5}{*}{A:INT8,} & SQ+ & 39.51 & 70.50 & 79.33 & 51.85 & 29.60 & 74.92 & 47.08 & 66.61 & 57.42 \\
    \multirow{-11}{*}{W:4}  & \multirow{-5}{*}{W:INT4} & AWQ+ & 46.76 & \textbf{78.16} & 81.13 & \textbf{56.11} & 30.80 & 77.97 & 47.59 & \textbf{71.59} & 61.26 \\ \hline
                       &            & RTN  & 27.39 & 51.35 & 61.50 & 28.44 & 24.40 & 56.47 & 40.58 & 52.41 & 42.82 \\ 
                       &            & GPTQ & 35.24 & 63.09 & 64.31 & 51.07 & 28.40 & 74.32 & 44.17 & 64.17 & 53.09 \\
                       &            & SQ   & 26.88 & 52.06 & 62.17 & 38.88 & 23.20 & 65.61 & 38.38 & 56.51 & 45.46 \\
                       &            & AWQ  & \textbf{43.34} & 72.81 & 69.51 & 52.82 & \textbf{29.00} & 75.63 & 45.39 & 62.27 & 56.35 \\
                       & \multirow{-5}{*}{A:MXINT8,} & SQ+ & 29.44 & 55.18 & 63.33 & 45.79 & 23.40 & 67.95 & 37.92 & 62.35 & 48.17 \\
                       & \multirow{-5}{*}{W:MXINT3} & AWQ+ & 42.49 & \textbf{75.13} & \textbf{77.55} & \textbf{53.28} & \textbf{29.00} & \textbf{76.50} & \textbf{46.06} & \textbf{67.64} & \textbf{58.46} \\ \cdashline{2-12}
                       &            & RTN  & 21.76 & 25.51 & 53.30 & 25.20 & 15.20 & 51.36 & 32.86 & 50.12 & 34.41 \\ 
                       &            & GPTQ & 22.70 & 39.69 & 56.21 & 34.68 & 15.40 & 57.56 & 35.47 & 50.51 & 39.03 \\
                       &            & SQ   & 21.76 & 25.38 & 41.65 & 25.83 & 15.40 & 52.56 & 32.65 & 50.83 & 33.26 \\
                       &            & AWQ  & \textbf{32.59} & 55.51 & 63.12 & 39.66 & \textbf{24.00} & 66.70 & 39.56 & 58.96 & 47.51 \\
    \multirow{-11}{*}{A:8,} & \multirow{-5}{*}{A:INT8,} & SQ+ & 21.93 & 25.63 & 40.46 & 26.13 & 14.80 & 51.36 & 34.49 & 49.64 & 33.06 \\
    \multirow{-11}{*}{W:3}  & \multirow{-5}{*}{W:INT3} & AWQ+ & 31.14 & \textbf{57.66} & \textbf{65.32} & \textbf{42.65} & 23.80 & \textbf{68.28} & \textbf{43.81} & \textbf{62.75} & \textbf{49.43} \\
    \bottomrule

\end{tabular}
\label{tab:Qwen_eval_harness}
\end{table*}

\end{document}